\documentclass{bmvc2k}
\usepackage{bbding}

\def\etal{\emph{et al}\bmvaOneDot}
\usepackage{textcomp}
\usepackage{stfloats}
\usepackage{url}
\usepackage{verbatim}
\usepackage{color}
\usepackage{amsmath,amsfonts}
\usepackage{algorithmic}
\usepackage{algorithm}
\usepackage{array}
\usepackage{graphicx}
\usepackage{subcaption}
\usepackage{booktabs}
\usepackage{nicefrac}
\usepackage{microtype}
\usepackage{xcolor}
\usepackage{multirow}
\usepackage{multicol}

\usepackage[T1]{fontenc}
\usepackage{hyperref}
\definecolor{MocoGreen}{RGB}{57,181,74}

\usepackage{amsmath,amsfonts,bm}

\def\eqref#1{equation~\ref{#1}}

\def\1{\bm{1}}

\DeclareMathAlphabet{\mathsfit}{\encodingdefault}{\sfdefault}{m}{sl}
\SetMathAlphabet{\mathsfit}{bold}{\encodingdefault}{\sfdefault}{bx}{n}

\begin{document}
\title{Topology-preserving Adversarial Training for Alleviating Natural Accuracy Degradation}
\addauthor{Xiaoyue~Mi}{mixiaoyue19s@ict.ac.cn}{1,2}
\addauthor{Fan~Tang$^\ast$}{tfan.108@gmail.com}{1,2}
\addauthor{Yepeng~Weng}{wengyepeng15@mails.ucas.ac.cn}{1,2}
\addauthor{Danding~Wang}{wangdanding@ict.ac.cn}{1,2}
\addauthor{Juan~Cao}{caojuan@ict.ac.cn}{1,2}
\addauthor{Sheng~Tang}{ts@ict.ac.cn}{1,2}
\addauthor{Peng~Li$^\ast$}{lipeng@air.tsinghua.edu.cn}{3}
\addauthor{Yang~Liu}{liuyang2011@tsinghua.edu.cn}{3,4}
\addinstitution{Institute of Computing Technology, Chinese Academy of Sciences (CAS), Beijing, China}
\addinstitution{University of Chinese Academy of Sciences, Beijing, China}
\addinstitution{Institute for AI Industry Research (AIR), Tsinghua University, Beijing, China}
\addinstitution{Department of Computer Science \& Technology, Tsinghua University, Beijing, China}
\runninghead{Mi, et al}{Topology-Preserving Adversarial Training}
\maketitle
\begin{abstract}
Despite the effectiveness in improving the robustness of neural networks, adversarial training has suffered from the natural accuracy degradation problem, i.e., accuracy on natural samples has reduced significantly.
In this study, we reveal that natural accuracy degradation is highly related to the disruption of the natural sample topology in the representation space by quantitative and qualitative experiments.
Based on this observation, we propose Topology-pReserving Adversarial traINing (TRAIN) to alleviate the problem by preserving the topology structure of natural samples from a standard model trained only on natural samples during adversarial training.
As an additional regularization, our method can be combined with various popular adversarial training algorithms, taking advantage of both sides.
Extensive experiments on CIFAR-10, CIFAR-100, and Tiny ImageNet show that our proposed method achieves consistent and significant improvements over various strong baselines in most cases.
Specifically, without additional data, TRAIN achieves up to $\mathbf{8.86\%}$ improvement in natural accuracy and $\mathbf{6.33\%}$ improvement in robust accuracy. 
\end{abstract}

\vspace{-2mm}
\section{Introduction}\label{Indro}

Adversarial training~\citep{schott2018towards,pang2022robustness,HE2023706} has been proven to effectively defense adversarial attacks~\cite{madry2017towards,GAO2023516,LI2023925} of neural networks~\citep{athalye2018obfuscated,tramer2020adaptive}.
However, models trained by adversarial training strategy have shown a significant reduction of accuracy in natural samples~\citep{madry2017towards}, which is usually called {\it natural accuracy degradation}~\citep{cui2021learnable}.
This problem hinders the practical application of adversarial training, as natural samples are the vast majority in reality~\citep{10.1145/3533767.3534373}.

Existing works attempt to alleviate natural accuracy degradation by data augmentation or extra data collection~\citep{rebuffi2021data,pmlr-v202-wang23ad}, distilling classifier boundary of the standard model~\citep{arani2020adversarial,cui2021learnable, chen2021ltd}, instance reweighting~\citep{zhang2021geometry}, early-stopping~\citep{zhang2020attacks}, adjustments of loss functions~\citep{pang2022robustness}, and learnable attack strategies during training~\citep{jia2022adversarial,kuurila2023adaptive}.
Nevertheless, these approaches have not fully closed the natural accuracy gap between adversarial and standard training.

\begin{figure*}[!t]
\centering
\begin{minipage}{0.25\textwidth}
  \centering
  \includegraphics[width=\textwidth]{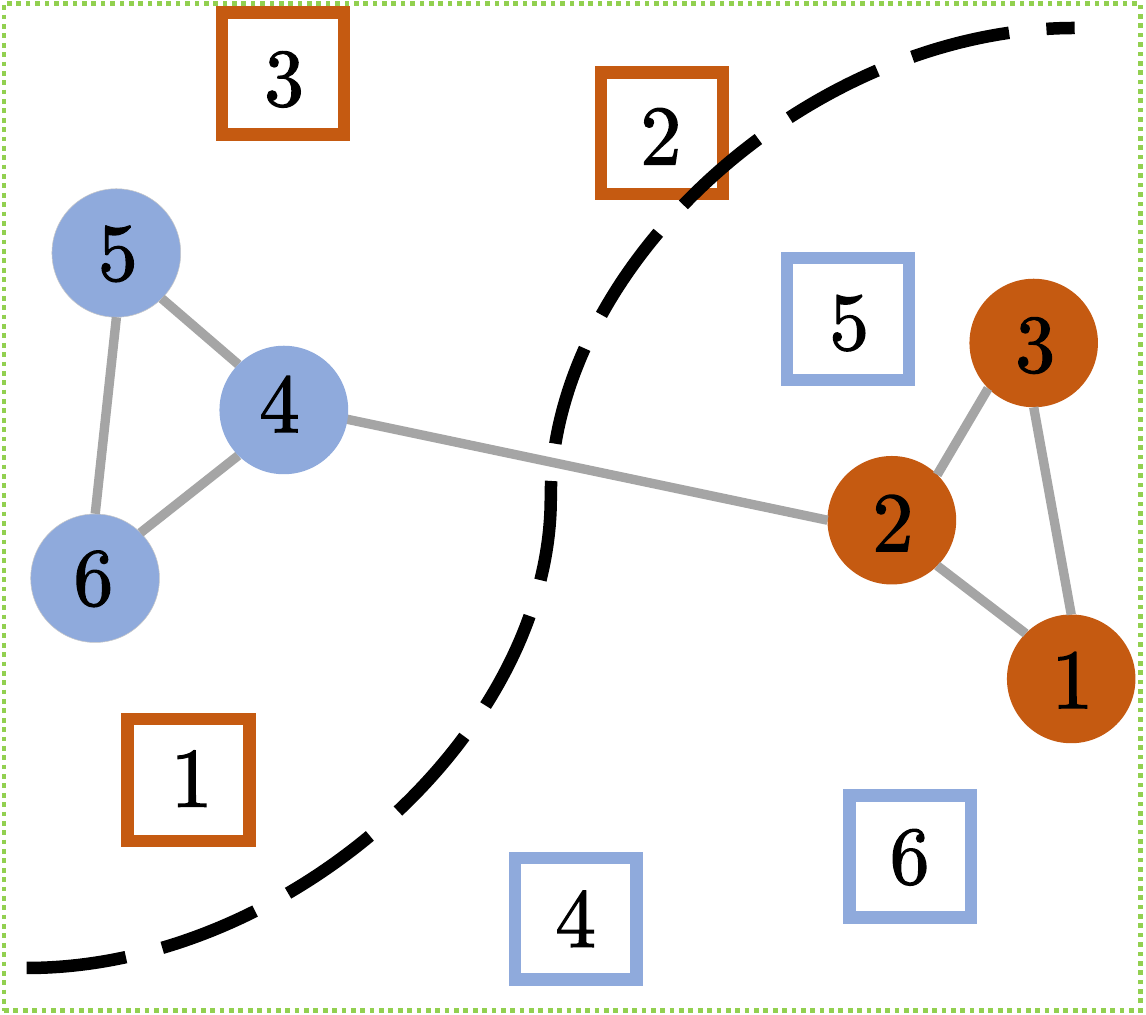}
  \vspace{1mm}
  \caption*{(a)Standard training}
  \label{fig:st}
\end{minipage}%
\hfil
\begin{minipage}{0.25\textwidth}
  \centering
  \includegraphics[width=\textwidth]{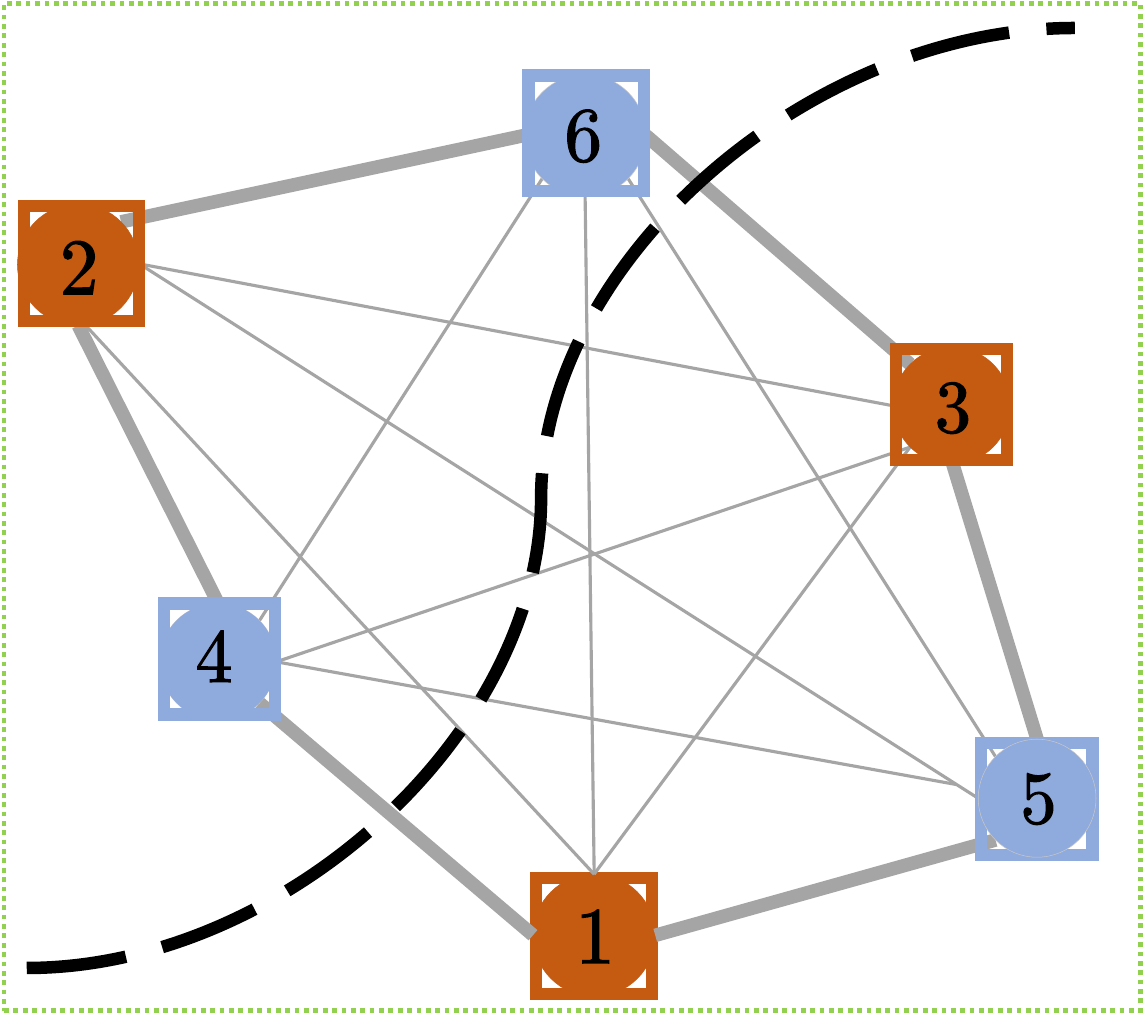}
  \vspace{1mm}
  \caption*{(b)Adversarial training}
  \label{fig:at}
\end{minipage}%
\hfil
\begin{minipage}{0.25\textwidth}
  \centering
  \includegraphics[width=\textwidth]{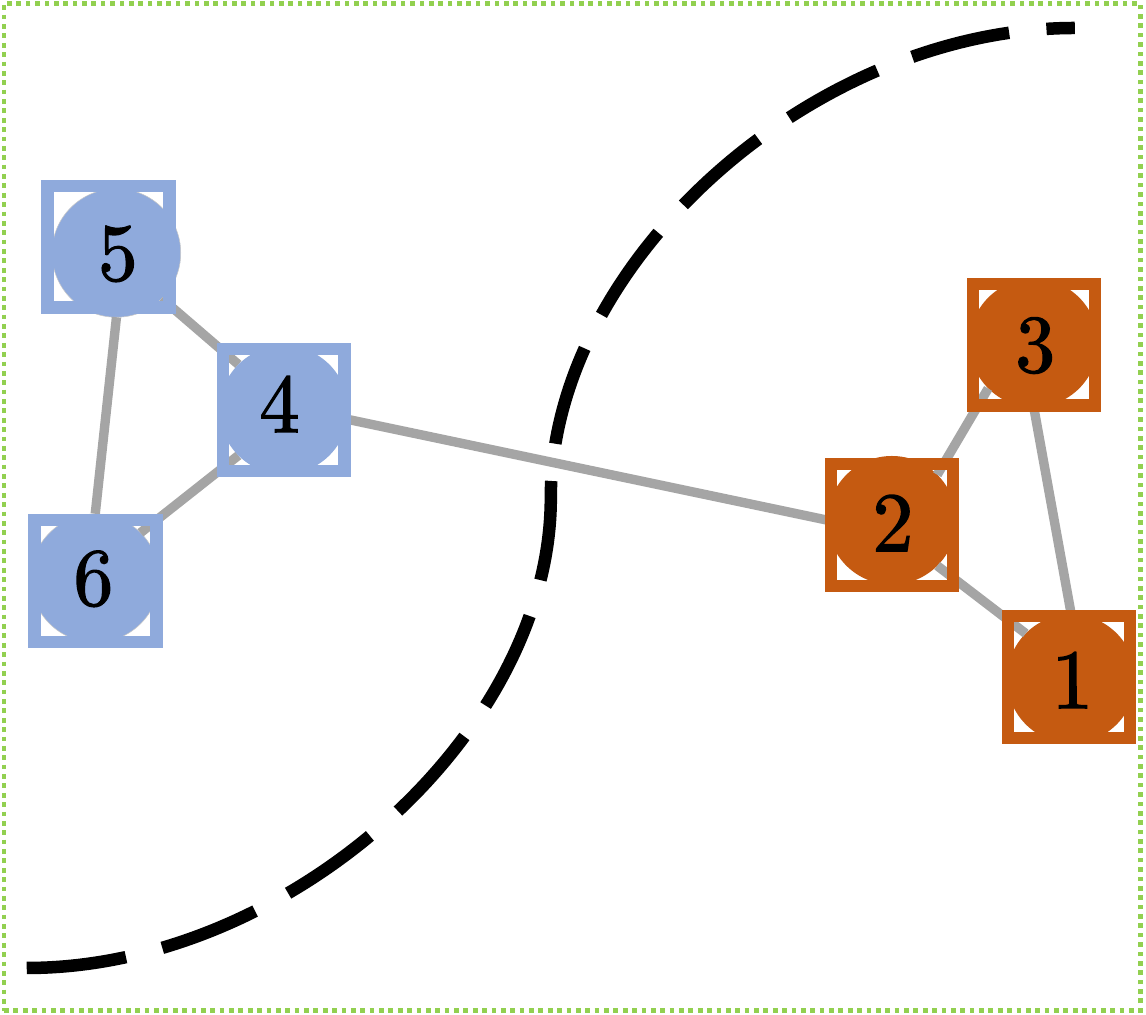}
  \vspace{1mm}
  \caption*{(c)TRAIN}
  \label{fig:or}
\end{minipage}%
\hfil
\begin{minipage}{0.25\textwidth}
  \centering
  \includegraphics[width=\textwidth]{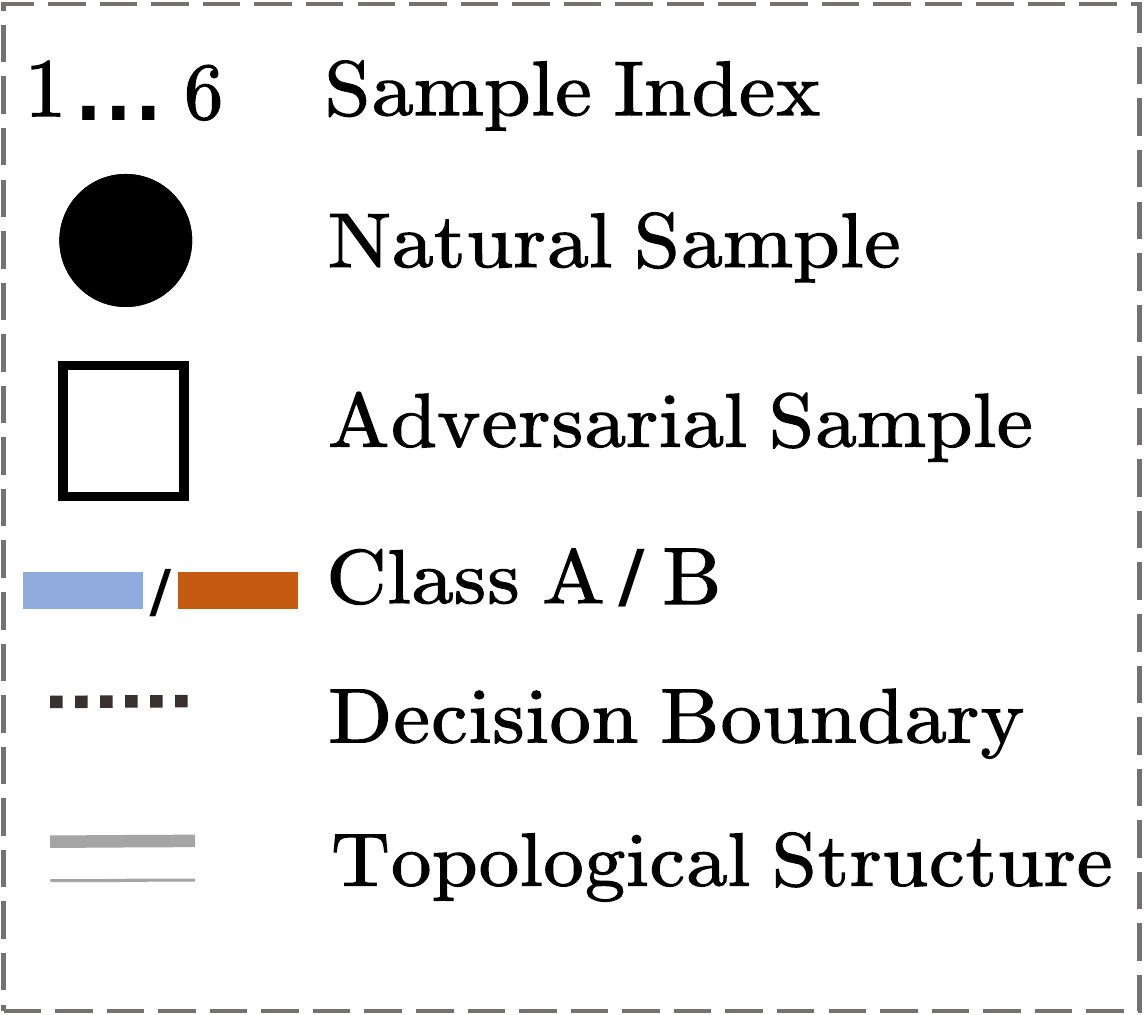}
  \vspace{1mm}
  \caption*{}
  \label{fig:tuli}
\end{minipage}
\vspace{1mm}
\caption{Illustrations for representation space under different training strategies.}
\vspace{-4mm}
\label{fig:insight}
\end{figure*}

Unlike previous efforts, we attempt to explain the natural accuracy degradation from a new perspective, topology.
Topology refers to the neighborhood relation of data in the representation space~\cite{rabunal2009encyclopedia}.
Some adversarial training studies~\cite{yang2021structure,qian2021improving} have shown the importance of topology in adversarial robustness generalization.
However, they do not attenuate the negative impact on the natural samples produced by the adversarial samples, resulting in incomplete topology preservation, the natural accuracy degradation still exists.

We conjecture that adversarial training destroys the topology of natural samples in the representation space, leading to a decrease in natural accuracy.
As illustrated in Fig.~\ref{fig:insight}(a), a model after standard training has a well-generalizing topology of natural samples but is vulnerable to adversarial samples, which are usually far from their true class distribution in the representation space.
Adversarial training pulls simultaneously the adversarial samples and their corresponding natural samples nearer~\citep{mao2019metric}~(Fig.~\ref{fig:insight}(b)) to improve the robustness of the model while leading to the poor topological structure of the natural sample features due to the negative influence of the adversarial samples.
Qualitative and quantitative analyses support the intuition that natural accuracy correlates with the topology preservation extent (see Sec.~\ref{Empirical_analysis} for more details).

Inspired by the above intuition, we propose a new approach called Topology-pReserving Adversarial traINing (TRAIN) to alleviate natural accuracy degradation~(Fig.~\ref{fig:insight}(c)), which closes the gap between adversarial and corresponding natural samples while preserving the well-generalizing topology of the standard model.
A straightforward solution is to distill the natural sample features of the standard model or the relationships based on the absolute distance between samples during adversarial training.
However, it suffers from optimization difficulties due to the great gap between standard and adversarial models.
So we construct the topological structure of data in the representation space based on the neighbor graph for each model.
We define the edge weight of the graph as the probability that different samples are neighbors, and topology preservation is achieved by aligning the standard model's graph and the adversarial model's graph.
Meanwhile, the optimization process of the standard model is not affected by the adversarial model, to reduce the negative impact of adversarial samples.
Experiments show that benefitting from topology preservation, TRAIN improves both the natural and robust accuracy when combined with other adversarial training algorithms.

Our contributions are as follows:
\vspace{-1.8mm}
\begin{itemize}
    \item[$\bullet$] We reveal that the topology of natural samples in the representation space plays an important role in the natural accuracy of adversarial models, which provides a new perspective on mitigating natural accuracy degradation.
\vspace{-1.5mm}
  \item[$\bullet$] We propose a {\it topology preservation adversarial training} method that preserves the topology structure between natural samples in the standard model representation space, which can be combined with various adversarial training methods.
\vspace{-1.5mm}
  \item[$\bullet$] Extensive quantitative and qualitative experiments on CIFAR-10, CIFAR-100, and Tiny ImageNet datasets show the effectiveness of the proposed TRAIN (maximum $8.86\%$ improvement for the natural accuracy and $6.33\%$ for the robust accuracy).
\end{itemize}
\vspace{-6.8mm}
\section{Related Work}
\label{related_work}
\vspace{-1mm}
\subsection{Adversarial Training}
Adversarial training~\citep{madry2017towards,zhang2019theoretically,wang2019improving,wong2020fast,wu2020adversarial,li2021subspace,jia2022adversarial,singh2023revisiting} is a prevailing method to improve the adversarial robustness of DNNs.
However, it decreases the accuracy of natural samples while increasing the adversarial robustness compared with standard training.
This phenomenon is called ``natural accuracy degradation'' or ``the trade-off between robustness and accuracy''.
Several works have been proposed to alleviate this problem.
Zhang~\etal~\cite{zhang2020attacks} used early-stopping.
Rebuffi~\etal~\citep{rebuffi2021data} tried to use more training data by data augmentation or adding extra data.
Researchers~\citep{arani2020adversarial,cui2021learnable, chen2021ltd} tried to distill the natural sample logits from the standard model to the adversarial model.
Zhang~\etal~\citep{zhang2021geometry} made use of instance reweighting.
Pang~\etal~\cite{pang2022robustness} redefined adversarial training optimization goals.
And Jia~\etal~\citep{jia2022adversarial} used reinforcement learning to obtain learnable attack strategies.
Different from them, we mitigate this problem from the view of the topology of different data in the representation space.
Some works~\cite{yang2021structure,qian2021improving} show topology is crucial for adversarial robustness generalization but ignore the negative impact of adversarial examples, and still degrade in natural accuracy.

\subsection{Knowledge Distillation in Adversarial Training}
Knowledge distillation can transfer knowledge from a larger, cumbersome model (teacher) to a smaller, more efficient model (student), which is commonly used for model compression.
Recently some algorithms have applied knowledge distillation to adversarial training.
Some works~\cite {goldblum2020adversarially,zi2021revisiting} distilled large robust models for robust model compression.
Different from them, researchers~\citep{arani2020adversarial,cui2021learnable, chen2021ltd} distilled the natural data logits of the standard model to enhance adversarial training on natural accuracy.
\citep{chen2021ltd} considered additional temperature factors during distillation.
However, they did not constrain the topology of samples in the representation space, and their distillation loss updates both standard and adversarial models simultaneously. Therefore, they were still negatively affected by adversarial examples.
The experimental section also includes comparative evaluations of different knowledge distillation methods.
\section{Topology's Role in Adversarial Training}
\label{headings}

\subsection{Formulation}

Following vanilla AT~\citep{madry2017towards}, the goal of adversarial training is defined as:
\begin{equation}
\mathop{\arg\min} _{\theta} \mathbb{E}_{(x, y) \in D}\left(\max_{\delta \in S} {L}(x+\delta, y;\theta)\right),
\label{eq:formulate}
\end{equation}
where $D$ is the data distribution for input $x$ and its corresponding label $y$, $\theta$ is the model parameters. 
$\delta$ stands for the perturbation applied to $x$ and is usually limited by perturbation size $\epsilon$.
$S =\left \{ \delta | \left \| \delta \right \|_{p} \le \epsilon  \right \}$ is the feasible domain for $\delta$.
$L\left( \cdot \right)$ usually is the cross-entropy loss for classification. 
By min-max gaming, adversarial training aims to correctly recognize all adversarial examples (${x'} = x+\delta$).
For descriptive purposes, we refer to models trained only on natural samples as {\it standard models} and those trained using adversarial training as {\it adversarial models} in the latter part.



\begin{figure*}[!t]
\centering
\begin{minipage}[b]{0.32\textwidth}
    \centering
    \includegraphics[width=\textwidth]{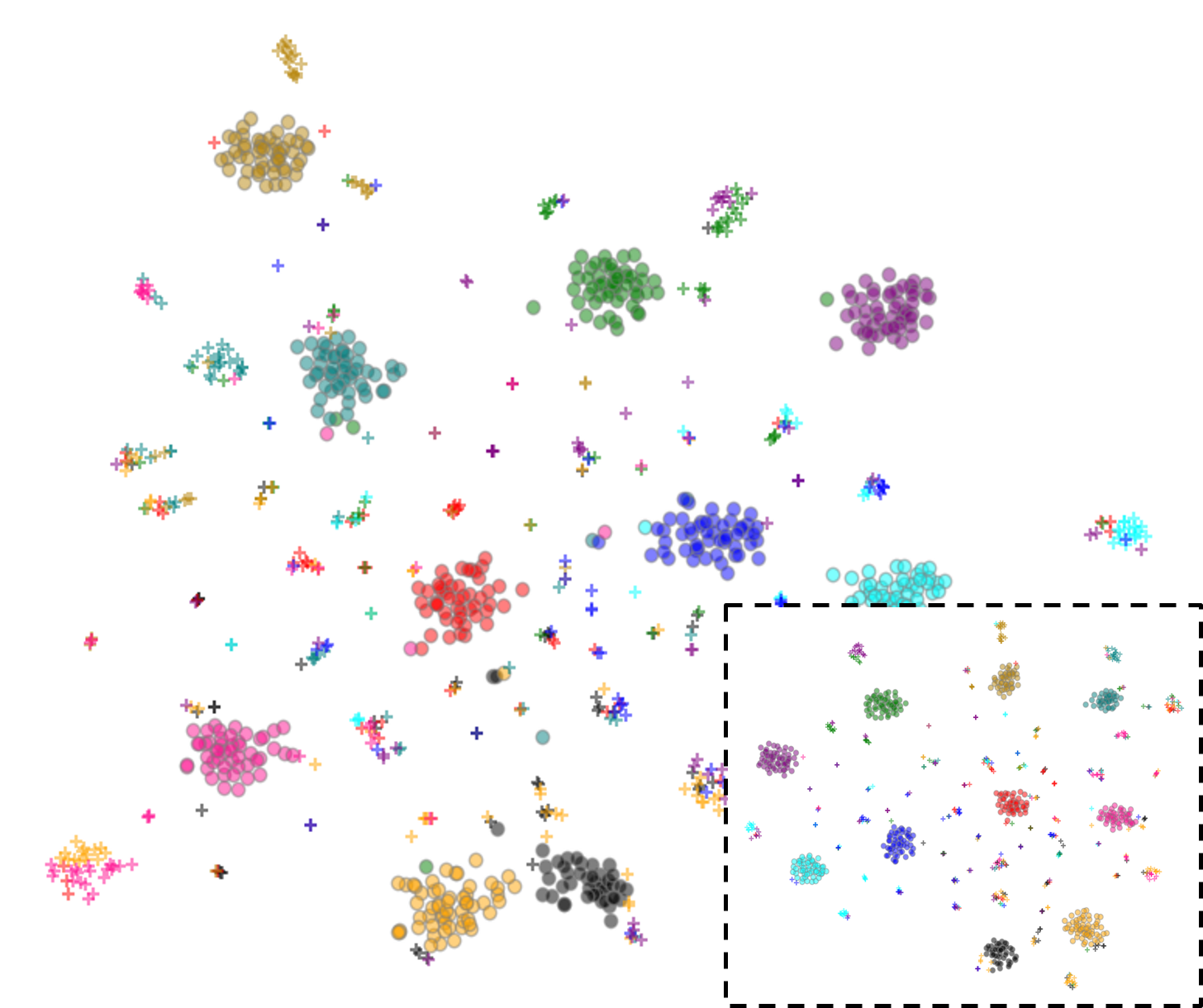}
    \vspace{1mm}
    \caption*{Standard (\textcolor{red}{77.39\%}/\textcolor{blue}{0.00\%})}
    \label{fig:vis-st-cifar100}
\end{minipage}
\hfill
\begin{minipage}[b]{0.32\textwidth}
    \centering
    \includegraphics[width=\textwidth]{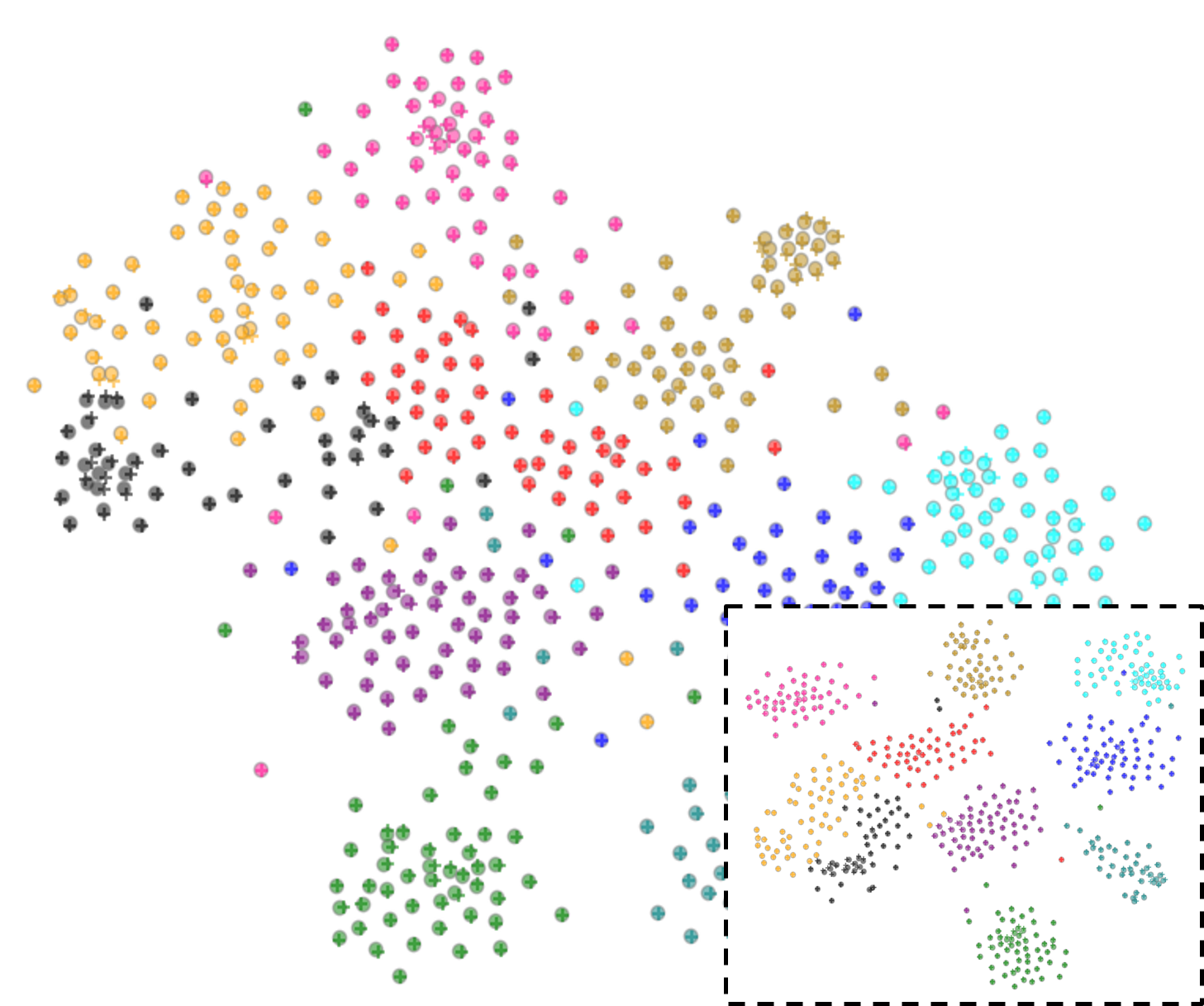}
    \vspace{1mm}
    \caption*{TRADES (\textcolor{red}{56.50\%}/\textcolor{blue}{30.93\%})}
    \label{fig:vis-at-cifar100}
\end{minipage}
\hfill
\begin{minipage}[b]{0.31\textwidth}
    \centering
    \includegraphics[width=\textwidth]{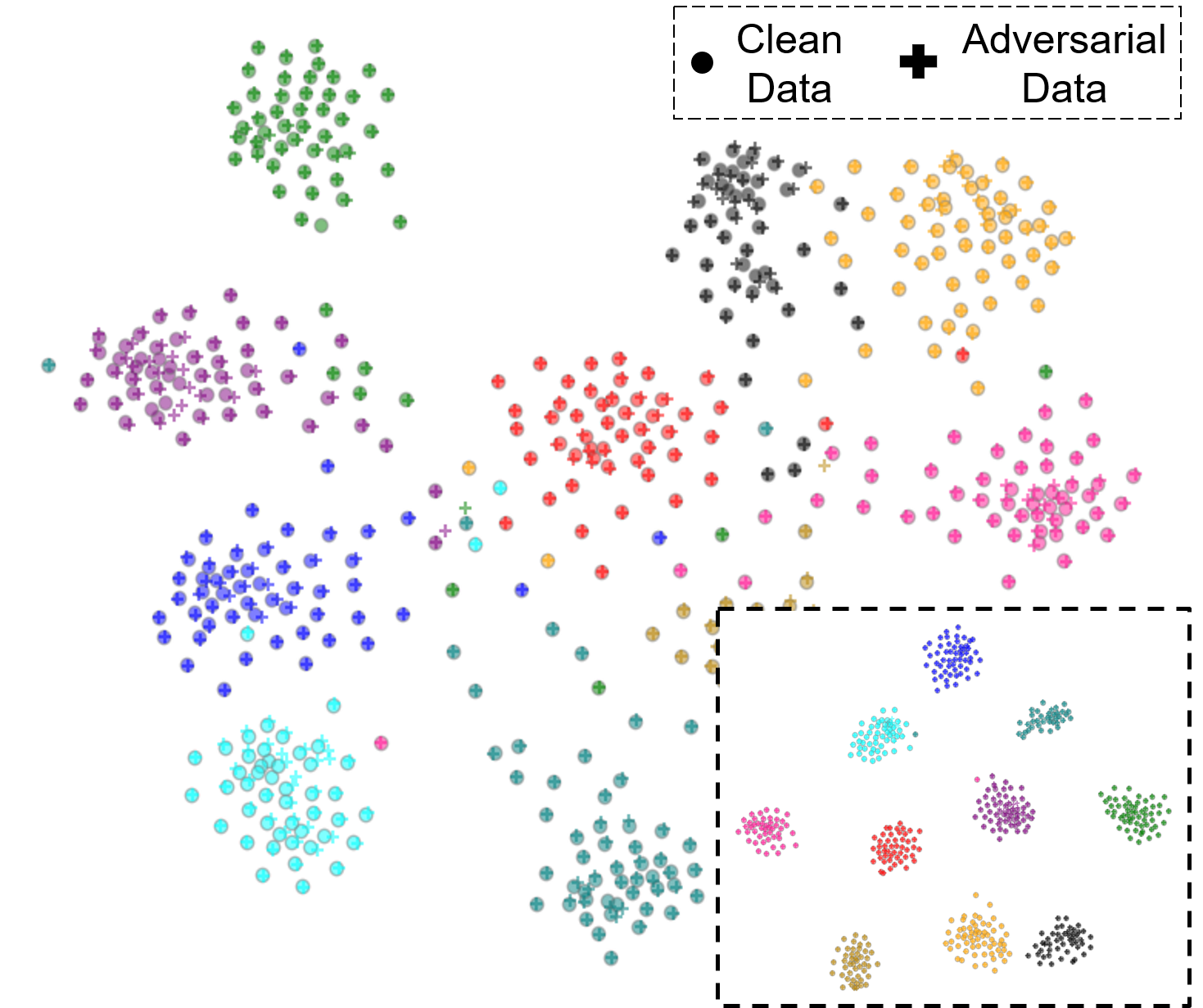}
    \vspace{1mm}
    \caption*{TRAIN (\textcolor{red}{65.28\%}/\textcolor{blue}{33.97\%})}
    \label{fig:vis-ours-cifar100}
\end{minipage}
\vspace{2mm}
\caption{
Analytical experiments reveal the relationship between topology quality in the representation space and natural accuracy.
(a), (b), and (c) show the differences in the representation space for the standard model, adversarial model (trained by TRADES with $\beta=6.0$), and TRAIN on CIFAR-100 training (small plots), and test sets (large plots).
Natural accuracy and PGD-20 accuracy are indicated in \textcolor{red}{red} and \textcolor{blue}{blue}, respectively.
}
\vspace{-4mm}
\label{Figure2}
\end{figure*}

\begin{figure}[!t]
\centering
\includegraphics[width=0.3\textwidth]{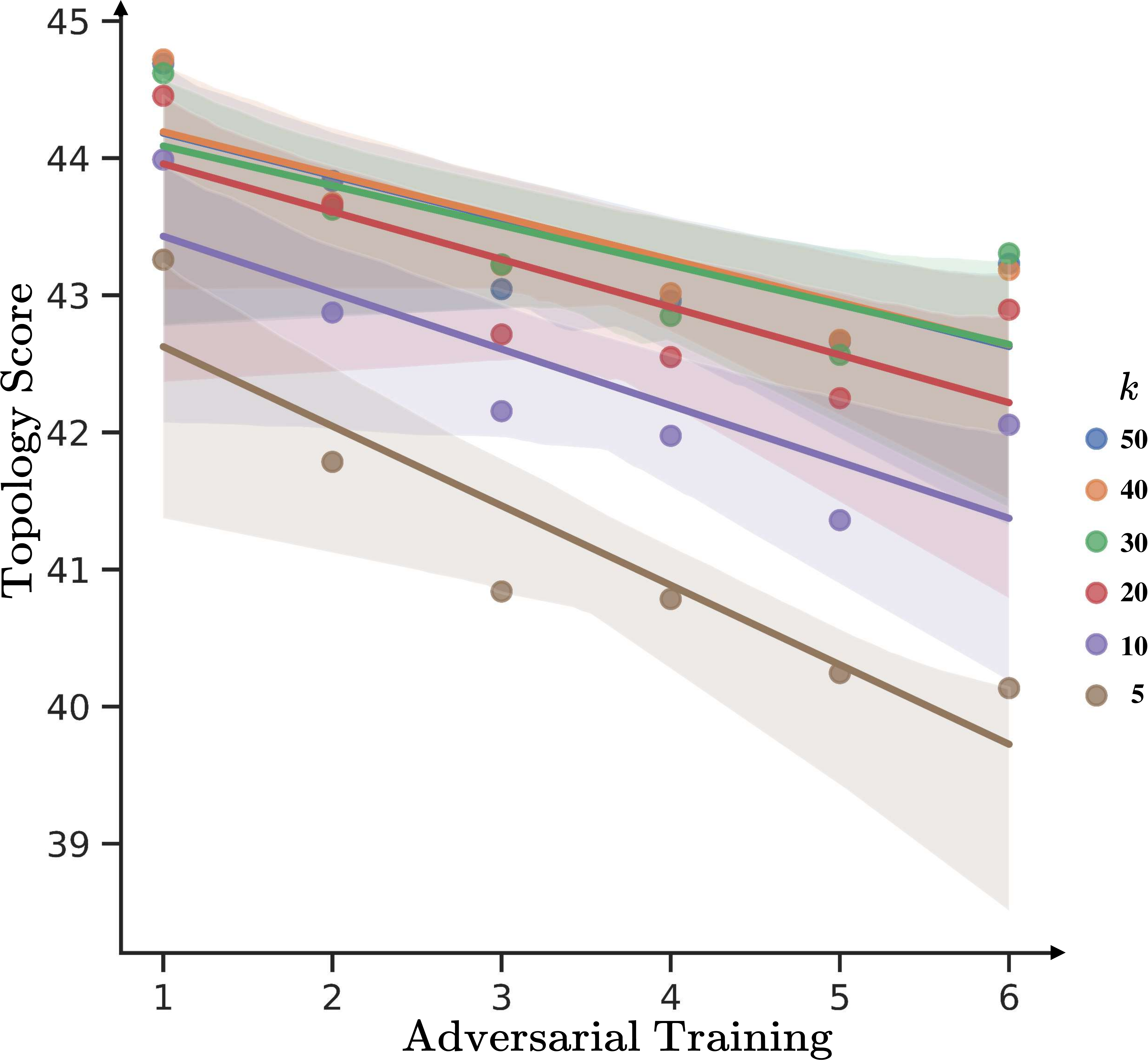}
\vspace{2mm}
\caption{Quantitative analysis reveals a negative correlation between the adversarial strength and the topology score.}
\vspace{-5mm}
\label{negative_correlation}
\end{figure}
\subsection{Empirical Analysis}
\label{Empirical_analysis}
In this section, we analyze how adversarial training influences topological relationships compared with standard models.
We find that the quality of the topology is positively correlated with natural accuracy, while negatively correlated with adversarial strength.
Adversarial models are trained by TRADES~\citep{zhang2019theoretically}, and here we consider the weights of the adversarial loss function $\beta$ as adversarial strength.
Larger $\beta$ represents the greater strength of adversarial training.
We choose the penultimate layer representations (before logits) of the standard model and adversarial models for qualitative and quantitative experimental analysis.
\textbf{See supplementary material for more details on experimental settings.}

\textbf{Qualitative analysis.}
As shown in Fig.~\ref{Figure2}, compared with the standard model on both training the test sets, the representation visualization for adversarial models shows more robustness, but a worse topology of data resulting in lower discrimination in different classes.

\textbf{Quantitative analysis.}
We conduct quantitative analysis by setting the $\beta = 1,2,...,6$ for TRADES as different adversarial strengths, and use $k$NN accuracy as the {\it topology score} to evaluate the quality of topology for different models, which is often used in manifold learning \citep{van2008visualizing,mcinnes2018umap} to evaluate the quality of topology in dimension reduction.
The higher the score, the more reasonable topology between the samples.
Specifically, we use both natural and adversarial data (generated by PGD-$20$) in the CIFAR-100 training set as the support set to predict the labels of natural and adversarial data in the test set.
To verify the reliability of the observation conclusion, we choose $k = 5,10,20,30,40,50$, respectively.

Fig.~\ref{negative_correlation} shows the strength of adversarial training and their corresponding topology qualities for different $k$.
A negative correlation between the strength of adversarial training and the topology quality could be observed.

\textbf{Why does adversarial training destroy topological relationships?} Adversarial representations are usually far away from their true class distribution, while natural samples are not.
Adversarial training narrowing the adversarial representations and natural representations concurrently usually makes the representation of natural samples further away from the original distribution, and hurts the topology and discrimination of natural data representations.
Zhang~\etal~\citep{zhang2021geometry} points out that adversarial training is equivalent to a special kind of regularization and has a strong smoothing effect, which also supports our intuition. 

\section{Topology-Preserving Adversarial Training}
\label{sec:frpkt}
\subsection{Overall Framework}
\label{overall}
To reduce the negative impact of adversarial samples, we propose a method TRAIN that focuses on preserving the topology of natural features from the standard model during adversarial training.
As shown in Alg.~\ref{Alg:1}, we train two models simultaneously: a standard model $M$ with a cross-entropy loss $L_{\rm ST}(\cdot)$ and an adversarial model ${M'}$ which is updated by a specific adversarial training algorithm.
For natural sample $x_{i}$, the outputs of $M(x_{i})$ are the feature of the last layer $ f_{x_{i}}$ and logit ${logit}_{x_{i}}$.
Similarly, the outputs of ${M'}({x_{i}'})$ are ${f'}_{{x_{i}'}}$ and ${{logit'}}_{{x_{i}'}}$ for aversarial sample ${x_{i}'}$ and ${f'}_{{x_{i}}}$ and ${{logit'}}_{x_{i}}$ for natural sample $x_{i}$.

The loss $L_{\rm ST}(\cdot)$ of ${M}$ is formulated as:
\begin{equation}
 L_{\rm ST} = L(z_{x_{i}},y_i), z_{x_{i}}=\frac{exp(logit_{{x}_{i}})}{\sum_{j=1}^N exp(logit_{{x}_{j}})},
\label{eq:st} 
\end{equation}
where ${L}$ is cross-entropy loss.
And the overall loss $L_{\rm AT}(\cdot)$ of ${M'}$ is formulated as follows:
\begin{equation}
 L_{\rm AT} = L_{\rm robust}({x'}) + \lambda L_{\rm TP}(M,{M'}),
\label{total} 
\end{equation}
where $L_{\rm robust}(\cdot)$ denotes the adversarial robustness loss, which is determined by the specific adversarial training algorithm employed. Additionally $L_{\rm TP}(\cdot)$ serves as a regularization item to preserve the topology of natural samples from $M$ and updates only ${M'}$.
A comprehensive discussion regarding the specifics of $L_{\rm TP}(\cdot)$ will be discussed in the next subsection.

\vspace{-3mm}
\subsection{Topology Preservation in Adversarial Training}
The topological structure is typically based on a neighborhood relation graph constructed by the similarity among samples in the representation space~\citep{rabunal2009encyclopedia,van2008visualizing,mcinnes2018umap}.
In this graph, each point is a sample in the representation space, while the edges are relationships among the samples, and the weights assigned to the edges are determined by the similarity between the samples.
Consequently, the topology preservation can be precisely formulated as follows:
\begin{equation}
L_{\rm TP} =
 \mathbb{E}_{(x, y) \in D}\left( {F}(P,Q)\right) \label{XY},
\end{equation}
where $P$ and $Q$ represent the neighborhood relation graph constructed by the inter-sample similarity for $M$ and ${M'}$, respectively.
$F(\cdot)$ measures the similarity between two graphs.

\textbf{Absolute relationship preservation.}
  Directly applying cosine similarity to calculate the pairwise distances $d_{ij}$ and ${d'}_{ij}$ between samples in representation spaces of $M$ and ${M'}$ to construct the neighborhood relation graph $P$ and $Q$ is a straightforward way:
\begin{equation}
P = \{d_{ij}| 0 < i,j \le N\}, Q = \{{d}_{ij}'| 0 < i,j \le N\},
\end{equation}
where $d_{ij}$ and ${d'}_{ij}$ are defined as:
\begin{equation}
\displaystyle
d_{ij} = 1 - \frac{f_{x_i}^T f_{x_j}}{||f_{x_i} ||_2 ||f_{x_j} ||_2},  
\widetilde{d}_{ij}' = 1 - \frac{{f'}_{{x}_{i}'}^T {f'}_{{x}_{j}'}}{||{f'}_{{x}_{i}'} ||_2 ||{f'}_{{x}_{j}'} ||_2}.
\label{XZ}
\end{equation}

However, there exists a substantial difference in the representation space between the adversarial model and the standard model, making it challenging to optimize the preservation of direct absolute relationships.

\begin{algorithm}[]
    \caption{Topology-Preserving Adversarial Training}
    \label{Alg:1}
    \begin{algorithmic}[1] 
        \REQUIRE the step size of perturbations $\epsilon$, batch size $n$, learning rate $\alpha$, attack algorithm optimization iteration times $K$, the number of training epochs $T$, adversarial model ${M}'$ with its parameters ${\theta}'$, standard model $M$ with its parameters $\theta$, loss weight $\lambda$ and training dataset $(x,y) \in D$
        \ENSURE robust model ${M}'$ with ${\theta}'$
        \STATE {Randomly initialize $\theta$ ,  ${\theta}'$ 
        \FOR {$ i = 1, ... ,T $ }
            \STATE {Sampling a random mini-batch $X=\left \{ x_1,x_2,...,x_n \right \}$ and corresponding labels $Y=\left \{ y_1,y_2,...,y_n \right \}$  from $D$}
            \STATE{Generating adversarial data ${X}'=\left \{ {x}'_1, {x}'_2, ..., {x}'_n \right \}$ through attack algorithms (such as PGD-K, FGSM)  }
            \STATE{$f_{X},{\rm logit}_{X} = M(X)$}
            \STATE{${f'}_{{X'}},{\rm logit'}_{{X'}} = {M}'({X}')$}
             \STATE{Evaluate $ L_{\rm ST}$}~Eq.~(\ref{eq:st})
            \STATE{Evaluate $ L_{\rm AT} = \lambda L_{\rm TP} + L_{\rm robust} $}~Eq.~(\ref{total})
            \STATE{Update model parameters:}
                \STATE{$\theta = \theta -\alpha\frac{1}{n} {\textstyle \sum_{i=1}^{n}\nabla_{\theta }L_{\rm ST} }  $}
                 \STATE{${\theta}' = {\theta}' -\alpha\frac{1}{n} {\textstyle \sum_{i=1}^{n}\nabla_{{\theta}'}L_{\rm AT} }  $}
        \ENDFOR
    }
    \end{algorithmic}
\end{algorithm}
\textbf{Relative relationship preservation.} 
Considering the significant gap between standard and adversarial models, our objective is to use conditional probability distribution for modeling the relationships between samples.
Specifically, we define the edge weights of the neighborhood relation graph as the probability that distinct samples are neighbors, thus ensuring topology preservation through the alignment of the probability distributions of the two graphs.

Different from manifold learning~\citep{van2008visualizing,mcinnes2018umap} which uses the regular Kernel Density Estimation (KDE) for approximations of the conditional probabilities, we use the cosine similarity-based affinity metric.
This choice is motivated by the excessive hyper-parameter tuning requirements and unacceptable training costs associated with KDE in adversarial training.
\begin{equation}
    \begin{split}
        K_{cos}(f_{x_i},f_{x_j}) =& \frac{1}{2}\left( \frac{f_{x_i}^T f_{x_j}}{||f_{x_i} ||_2 ||f_{x_j} ||_2}+1 \right), \\
                           =&\frac{1}{2}(2-d_{ij}), \\
    \end{split}
\end{equation}
where $K_{cos}$ is cosine similarity-based
affinity metric value for $x_i$ and $x_j$.

Moreover, we add a special term $\rho_{j}$ to better preserve the global structure of representation space.
$\rho_{j}$ represents the distance from the $j_{th}$ data point to its nearest neighbor. 
Subtracting $\rho_{j}$ ensures the local connectivity of the graph, avoiding isolated points and thus better preserves the global structure. 
\begin{equation}
\widetilde{d}_{ij} =  d_{ij}-\rho_{j},
~\widetilde{d}_{ij}' =  {d}_{ij}'-{\rho_{j}'}.
\end{equation}

After normalization, we obtain the $p_{i|j}$, which represents the conditional probability that the $i_{th}$ natural sample is a neighbor of the $j_{th}$ natural sample in the representation space of $M$.
\begin{equation}
    \begin{split}
        p_{i|j} &= \frac{2-\widetilde{d}_{ij}}{\sum^{N}_{k =1,k\ne j} (2-\widetilde{d}_{jk})}. \\
    \end{split}
\end{equation}
Similarly, for the adversarial model~${M'}$:
\begin{equation}
    \begin{split}
        q_{i|j} &= \frac{2-\widetilde{d}_{ij}'}{\sum^{N}_{k =1,k\ne j} (2-\widetilde{d}_{jk}') }. \\
    \end{split}
\end{equation}
So the neighborhood relation graph construction of $M$ can be formalized as:
\begin{equation}
\displaystyle
 P = \left\{p_{i|j} \bigg| p_{i|j} = \frac{2-\widetilde{d}_{ij}}{\sum^{N}_{k =1,k\ne j} (2-\widetilde{d}_{jk}))}, 0 < i,j \le N\right\}.
\label{Xzz}
\end{equation}
Similarly, the relationship graph for ${M}'$ is: 
\begin{equation}
\displaystyle
Q = \left\{q_{i|j} \bigg|q_{i|j} = \frac{2-\widetilde{d}_{ij}'}{\sum^{N}_{k =1,k\ne j} (2-\widetilde{d}_{jk}') }, 0 < i,j \le N\right\}. 
\label{Xsz}
\end{equation}
We use cross-entropy loss to measure the similarity of $P$ and $Q$ for such flexible relationships.
Finally, the $L_{\rm TP}$ for TRAIN is formalized as: 
\begin{equation}
\begin{split}
L_{\rm{TP}} &= {CE}(P, Q) \\
&= \! \sum_{i} \! \! \sum_{j}\! \!\left[p_{i| j}\! \log \!\left(\frac{p_{i| j}}{q_{i| j}}\!\right)\! +\!\left(\!1-p_{i |j}\!\right)\! \!\log\! \!\left(\!\frac{1-p_{i| j}}{1-q_{i| j}}\!\right)\!\right]\!.
\label{eq:tp}
\end{split}
\end{equation}

\vspace{-6mm}
\section{Experiments}
\label{others}
\vspace{-3mm}
\textbf{Experimental settings}
Following~\citep{cui2021learnable,jia2022adversarial,pang2022robustness}, we conduct extensive evaluations on popular datasets in adversarial training, including CIFAR-10, CIFAR-100~\citep{krizhevsky2009learning}.
ResNet-18 is the backbone of standard models, and WideResNet-34-10 is the backbone of adversarial models.
The adopted adversarial attacking method during training is PGD-10, with a perturbation size $\epsilon = 0.031$, a step size of perturbations $\epsilon_1 = 0.007$. 
For different experiment settings, we choose different $\lambda$.
We set $\lambda = 5$ on CIFAR-10 dataset, and $\lambda = 20a$ on CIFAR-100 dataset, where $a = \frac{2}{1+ e^{-\frac{10t}{100}-1}}$ and $t$ is the current $t$-th epoch during training. 
Finally, all experiments were done on GeForce RTX 3090.

Our evaluation metrics are natural data accuracy (Natural Acc.) and robust accuracy (Robust Acc.).
Robust accuracy is the model classification accuracy under adversarial attacks.
Following previous works, we choose three representative adversarial attack methods for evaluation: PGD-20, C\&W-20~\citep{carlini2017towards}, and Auto Attack~\citep{croce2020reliable}. 
We denote the model's defense success rate under those attacks separately as \textit{PGD-20 Acc.}, \textit{C\&W-20 Acc.}, and \textit{AA Acc.}.
To provide a comprehensive evaluation and comparison with other state-of-the-art adversarial training methods, we use their original hyperparameters in our settings, and include baselines: Vanilia AT~\citep{madry2017towards}, TRADES~\citep{zhang2019theoretically},  LBGAT, MART~\citep{wang2019improving}, FAT~\citep{zhang2020attacks}, GAIRAT~\citep{zhang2021geometry}, AWP~\citep{wu2020adversarial}, SAT~\citep{sitawarin2021sat}, LAS~\citep{jia2022adversarial},
and ECAS~\citep{kuurila2023adaptive}.
For TRADES, we set $\beta = 6.0$. 
For LBGAT, we conduct experiments based on vanilla AT and TRADES ($\beta = 6.0$).
We also provide \textbf{details of experimental settings} and \textbf{experiments on Tiny ImageNet~\citep{deng2009imagenet} in supplementary materials}.

\vspace{-4mm}
\subsection{Main Results}
\vspace{-2mm}

\begin{figure*}[!t]
\centering
\begin{minipage}[b]{0.32\textwidth}
    \centering
    \includegraphics[width=\textwidth]{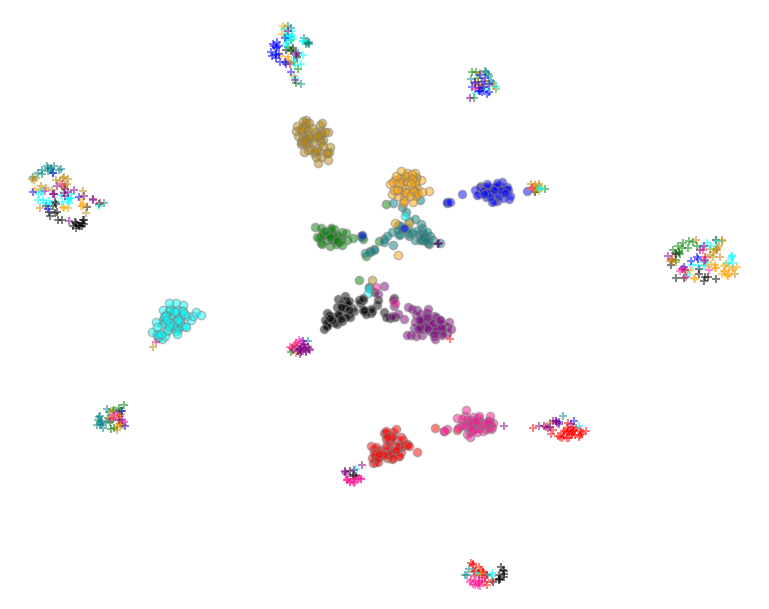}
    \vspace{1mm}
    \caption*{Standard training}
    \label{fig:vis_st}
\end{minipage}
\hfill
\begin{minipage}[b]{0.32\textwidth}
    \centering
    \includegraphics[width=\textwidth]{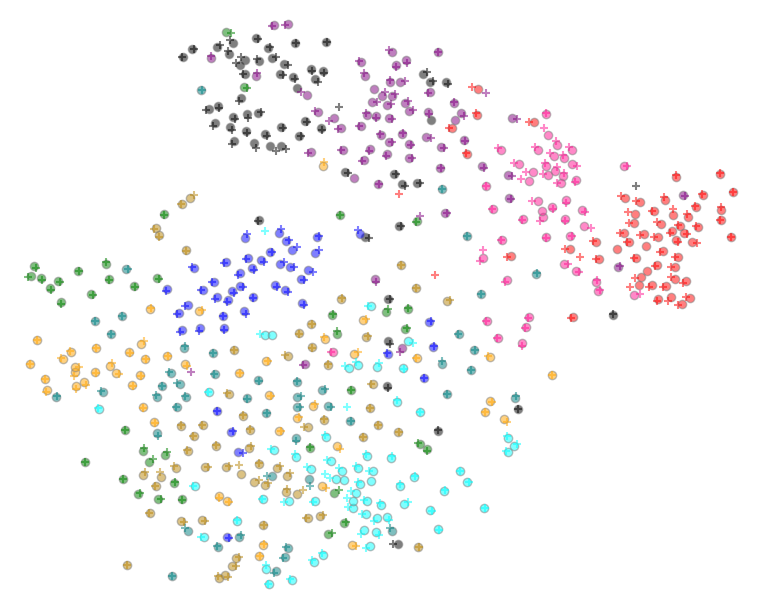}
    \vspace{1mm}
    \caption*{TRADES $(\beta=6.0)$}
    \label{fig:vis_trades}
\end{minipage}
\hfill
\begin{minipage}[b]{0.31\textwidth}
    \centering
    \includegraphics[width=\textwidth]{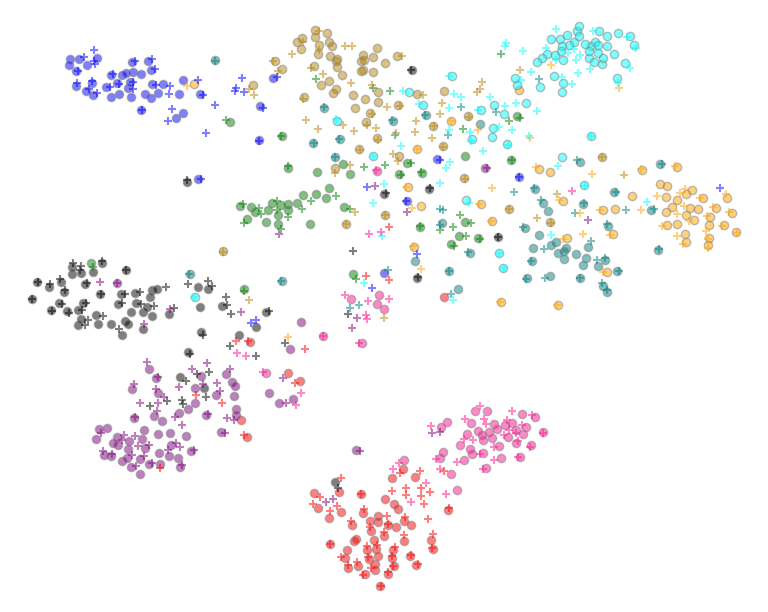}
    \vspace{1mm}
    \caption*{TRAIN}
    \label{fig:vis_trades_ours}
\end{minipage}
\vspace{2mm}
\caption{t-SNE visualizations of penultimate layer features on CIFAR-10. Crosses and circles are adversarial samples and natural samples, respectively. Different colors represent different classes.}
\vspace{-4mm}
\label{fig:tsne-cifar10}
\end{figure*}


\textbf{Quantitative results.}
As shown in Tables~\ref{NewCIFAR-10} and~\ref{NewCIFAR-100}, TRAIN achieves a better trade-off between natural accuracy and adversarial robustness compared with the most popular adversarial training algorithms.
TRADES, LBGAT, and ECAS achieve significant improvement in natural accuracy by combining with TRAIN, and the robust accuracy is also relatively improved or preserved.

\begin{table*}[t]
 
    \centering 
    \scalebox{0.6}{
    \begin{tabular}{lcccc}
    \toprule
    \multirow{2}{*}{\textbf{Defense}}
    &\multirow{2}{*}{\textbf{Natural Acc.}} &\multicolumn{3}{c}{\textbf{Robust Acc.}} \\
    ~&~&\textbf{PGD-20 Acc.}& \textbf{C\&W-20 Acc.}&\textbf{AA Acc.}\\
    \midrule 
    Vanilla AT*~\citep{madry2017towards} & 85.17 & 55.08  & 53.91 & 51.69 \\
    MART*~\citep{wang2019improving} & 84.17& \underline{58.56} & 54.58 & 51.10\\
    FAT* ~\citep{zhang2020attacks}& \textbf{87.97} & 49.86 & 48.65 & 47.48\\
    GAIRAT*~\citep{zhang2021geometry} & 86.30 & \textbf{59.54} & 45.57 & 40.30\\
    AWP*~\citep{wu2020adversarial} & 85.57 & 58.13 & 56.03 & 53.90\\
    \midrule
    ECAS~\citep{kuurila2023adaptive}& 84.57 & 55.86 & 54.65 & 52.10\\
    ECAS+TRAIN& 85.26\textcolor{MocoGreen}{$(\uparrow 0.69) $} & 56.23\textcolor{MocoGreen}{$(\uparrow 0.37) $} & 54.77\textcolor{MocoGreen}{$(\uparrow 0.12) $} & 52.22\textcolor{MocoGreen}{$(\uparrow 0.12) $}\\
    \midrule
    TRADES*~\citep{zhang2019theoretically} & 85.72 & 56.10 & 53.87 & 53.40\\
    LAS-TRADES*~\citep{jia2022adversarial} & 85.24 & 57.07 & 55.45 & 54.15\\
    TRADES + TRAIN & \underline{87.07}\textcolor{MocoGreen}{$(\uparrow 1.35) $} & 58.51\textcolor{MocoGreen}{$(\uparrow 2.41) $} & \textbf{56.81\textcolor{MocoGreen}{$(\uparrow 2.94) $}} & \textbf{54.70}\textcolor{MocoGreen}{$(\uparrow 1.30) $}\\
    \midrule
    TRADES + LBGAT~\citep{cui2021learnable} & 80.20 & 57.41 & 54.84 & 53.32\\
    TRADES + LBGAT +TRAIN& 86.69\textcolor{MocoGreen}{$(\uparrow 6.49) $} &58.04\textcolor{MocoGreen}{($\uparrow$0.63)} & \underline{56.75}\textcolor{MocoGreen}{$(\uparrow 1.91) $} & \underline{54.47}\textcolor{MocoGreen}{$(\uparrow 1.15) $} \\
\bottomrule
\end{tabular}}
\vspace{2mm}
\caption{Results on CIFAR-10. ``*'' are the results directly quoted from LAS.
The best and second best results are \textbf{bolded} and \underline{underlined}, respectively.
}
\label{NewCIFAR-10}
\vspace{-4mm}
\end{table*}
\begin{table*}[t]

    \centering 
    \scalebox{0.65}{
    \begin{tabular}{lcccc}
    \toprule
    \multirow{2}{*}{\textbf{Defense}}
    &\multirow{2}{*}{\textbf{Natural Acc.}} &\multicolumn{3}{c}{\textbf{Robust Acc.}} \\
    ~&~&\textbf{PGD-20 Acc.}& \textbf{C\&W-20 Acc.}&\textbf{AA Acc.}\\
    \midrule 
Vanilla AT*~\citep{madry2017towards} & 60.89 & 31.69  & 30.10 & 27.86 \\
    SAT*~\citep{sitawarin2021sat} & 62.82& 27.17 & 27.32 & 24.57\\
    AWP*~\citep{wu2020adversarial}& 60.38 & 33.86 & 31.12 & 28.86\\
     \midrule
    ECAS~\citep{kuurila2023adaptive}& 64.60 & 35.41 & \underline{33.39} & 29.55\\
    ECAS+TRAIN& 65.24\textcolor{MocoGreen}{$(\uparrow 0.64) $} & \textbf{35.83}\textcolor{MocoGreen}{$(\uparrow 0.42) $} & \textbf{33.50}\textcolor{MocoGreen}{$(\uparrow 0.11) $} & \textbf{30.69}\textcolor{MocoGreen}{$(\uparrow 1.14) $}\\
    \midrule
    TRADES*~\citep{zhang2019theoretically} & 58.61 & 28.66 & 27.05 & 25.94\\
    LAS-TRADES*~\citep{jia2022adversarial} & 60.62 & 32.53 & 29.51 & 28.12\\
    TRADES + TRAIN & \textbf{67.47}\textcolor{MocoGreen}{$(\uparrow 8.86) $} & 34.99\textcolor{MocoGreen}{$(\uparrow 6.33) $} & 31.61\textcolor{MocoGreen}{$(\uparrow 4.56) $} & 28.95\textcolor{MocoGreen}{$(\uparrow 3.01) $}\\
    
    \midrule
    TRADES + LBGAT*~\citep{cui2021learnable} & 60.64 & 34.75 & 30.65 & 29.33\\
    TRADES + LBGAT+ TRAIN & \underline{65.40}\textcolor{MocoGreen}{$(\uparrow 4.76) $} & \underline{35.46}\textcolor{MocoGreen}{$(\uparrow 0.71) $} & 32.36\textcolor{MocoGreen}{$(\uparrow 1.71) $} & \underline{30.17}\textcolor{MocoGreen}{$(\uparrow 0.84) $}\\
\bottomrule
\end{tabular}}
\vspace{2mm}
\caption{Results on CIFAR-100. ``*'' are the results directly quoted from LAS.
The best and second best results are marked in \textbf{bold} and \underline{underline}.
}
 \label{NewCIFAR-100}
\end{table*}

\textbf{Qualitative analysis.}
To showcase the efficacy of our algorithm in assisting the adversarial model in constructing a well-generalizing topology in the representation space, we use t-SNE to visualize samples from ten randomly selected categories in the CIFAR-100 test set and all categories of the CIFAR-10 test set for qualitative analysis. 
Figs.~\ref{Figure2} and~\ref{fig:tsne-cifar10} show the results of CIFAR-100/10 datasets, respectively.
For standard training (Figs.~\ref{fig:vis-st-cifar100} and~\ref{fig:vis_st}), the natural data exhibit clear clustering, while the adversarial samples appear disjointed, resulting in poor performance on robust accuracy.
The TRADES approach facilitates the alignment of natural and adversarial data to enhance robust accuracy. Nonetheless, it is noteworthy that this alignment process can unintentionally disrupt the integrity of natural feature topologies, as it lacks any defensive measures to counteract this effect (refer to Figs.~\ref{fig:vis-at-cifar100} and~\ref{fig:vis_trades} in the paper for visual representations of this phenomenon).
As shown in Figs.~\ref{fig:vis-ours-cifar100} and~\ref{fig:vis_trades_ours}, applying the proposed TRAIN to TRADES could drive the cluster for each category to be more compact, thereby preserving the topology more effectively.

\vspace{-4mm}
\subsection{Ablation Studies}
\vspace{-2mm}
\begin{table}
\centering 
    \resizebox{0.6\linewidth}{!}{
    \begin{tabular}{lcccc}
    \toprule
    \multirow{2}{*}{\textbf{Methods}} &\multirow{2}{*}{\textbf{Natural Acc}} &\multicolumn{3}{c}{\textbf{Robust Acc}} \\ 
    ~&~&\textbf{PGD-20 Acc}& \textbf{C\&W-20 Acc}&\textbf{AA Acc}\\
    \midrule 
    Vanilla AT~\citep{madry2017towards} & 60.44 & 28.06  & 27.85 & 24.81 \\
    MCA~\citep{yang2021structure} & 57.18 & 29.31 & 27.23 & 25.76 \\
    Vanilla AT + RKD~\citep{park2019relational}  & 64.00 & 28.32 & 27.92 & 24.92 \\
    Vanilla AT + CRD~\citep{tian2019contrastive}   & 62.22 & 27.47 & 27.42 & 24.53 \\
     Vanilla AT + TRAIN$'$  & 62.10 & 29.43 & 29.66  & 25.78\\
    Vanilla AT + TRAIN  & \textbf{66.39} & \textbf{29.88} & \textbf{29.84}  & \textbf{25.81}\\
    
\bottomrule
\end{tabular}
}
\vspace{2mm}
\caption{Ablation results on different relation-preserving methods.}
\label{Ablation1}
\end{table}
In this section, we delve into TRAIN to study its effectiveness in different relation-preserving methods
We present a comparative analysis of our proposed method with alternative approaches: a metric learning approach called MCA~\citep{yang2021structure} and two absolute relationship distillation methods, namely RKD~\citep{park2019relational} and CRD~\citep{tian2019contrastive}.
MCA applies a supervised contrastive loss into adversarial training. 
RKD takes the absolute value of the cosine distance between samples as the relationship as discussed in Sec.~\ref{sec:frpkt}.
 CRD requires that a sample's representation in the student model be closer to its corresponding representation in the teacher model while being farther from the representations of other samples in the teacher model.
TRAIN$'$ means adversarial training will influence the standard models during training.
Table~\ref{Ablation1} shows the effectiveness of the relative relationship preservation 
TRAIN$'$ means adversarial training will influence the standard models during training, and it is important to reduce the negative influence of adversarial samples (comparison between TRAIN and TRAIN$'$).
All the ablation experiments are based on the CIFAR-100 dataset and combined with TRADES, and we provide its other training details in the supplementary material.

\begin{table}[t]
    \vspace{2mm}
    \centering 
    \scalebox{0.75}{
    \begin{tabular}{ccc}
    \toprule
    \textbf{Vanilla AT} &  \textbf{Vanilla AT + LBGAT}  & \textbf{Vanilla AT + TRAIN}\\
    \midrule 
     821 & 848 & 849 \\
     \midrule
    \textbf{TRADES} & \textbf{TRADES + LBGAT} &  \textbf{TRADES + TRAIN}\\
    \midrule 
    1,079 & 1,106 & 1,109 \\
\bottomrule 
\end{tabular}
}
\vspace{2mm}
\caption{Training time in second of an epoch on one RTX 3090 GPU.}
    \label{tab:time_cost}
\end{table}

\textbf{Time complexity.}
Our method is based on batch computation, and its time complexity is $O(N({mz}'K)) +O(N(bz ({fz}'+fz) + mz)$, where ${mz}'$ and $mz$ is the number of neurons of the adversarial model ($48.32$ M) and standard model ($11.22$ M), $bz$ is the batch size ($128$), ${fz}'$ and $fz$ is the feature size of the standard model ($512$) and adversarial model ($640$), and $K$ is the number of iterations for generating adversarial examples ($10$). For classic adversarial training, its time complexity is $O(N({mz}'K)$. Since $bz ({fz}'+fz) + mz << {mz}'K$, the additional time overhead of our method is negligible.

Note that, the primary time-consuming factor in adversarial training algorithms lies in the need for additional backpropagation during the generation of adversarial samples, whereas the TRAIN algorithm does not incur any extra computational cost in this regard.
Table~\ref{tab:time_cost} shows the time statistics for training one epoch (with batch size equals $128$) by different baselines.
It takes an additional $28$ seconds when combined with Vanilia AT and $3\%$ ($30$ seconds) on TRADES for TRAIN, which is as fast as LBGAT.
We also analyze \textbf{the influence of batch size, hyper-parameter $\lambda$, and model architectures in supplementary material.}

\section{Conclusion}
\vspace{-2mm}
Compared with standard training, adversarial training shows significant natural accuracy degradation.
Different from previous algorithms, we assume this is due to topology disruption of natural features, and confirm it by empirical experiments.
Based on that, we propose Topology-pReserving Adversarial traINing (TRAIN).
While improving the adversarial robustness of the model, it preserves the topology of natural samples in the representation space of the standard model. 
Our method has been rigorously validated through both quantitative and qualitative experiments, demonstrating its effectiveness and reliability.

\section{Acknowledgments}
This work is supported by the National Natural Science Foundation of China (62203425).

\newpage
\section{Supplementary Material for Topology-preserving Adversarial Training for Alleviating Natural Accuracy Degradation}
\subsection{Pipeline}

\begin{figure*}
  \centering
  \includegraphics[width=0.5\linewidth]{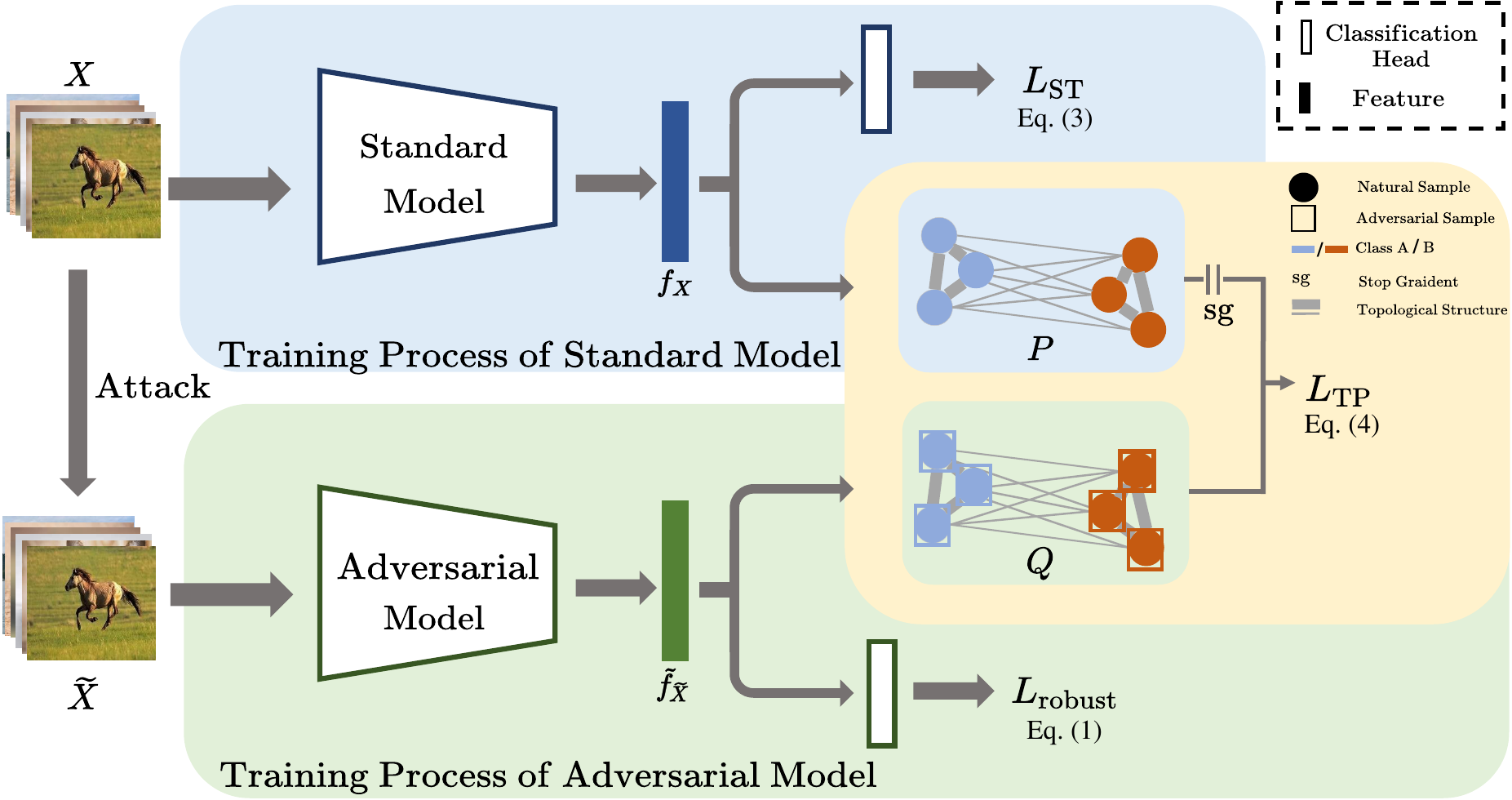}
  \vspace{2mm}
  \caption{Overall framework of TRAIN.
  Specifically, we train a standard model $M$ and an adversarial model ${M'}$. The standard model takes natural samples $X$ as input and is optimized by cross-entropy loss. On the other hand, the adversarial model takes adversarial samples ${X'}$ as input and is optimized by robust loss $L_{robust}(\cdot)$ and topology preservation loss $L_{\rm{TP}}(\cdot)$. $L_{\rm{TP}}(\cdot)$ constructs and aligns the neighborhood relation graph $P$ and $Q$ in the representation spaces of $M$ and ${M'}$, respectively.
  It can preserve the topological relationships among samples to reduce the negative effects of the adversarial samples during adversarial training.}
  \label{pipeline}
\end{figure*}

To make it easier for the reader to understand our method, Fig.~\ref{pipeline} shows the overall process for TRAIN combined with vanilla AT.

\textbf{Discussion.} LBGAT also adopts a two-model framework and transfers the prior knowledge of $M$ to ${M}'$.
Nonetheless, there exist notable distinctions between our proposed method and LBGAT, which can be summarized as follows:
\begin{enumerate}
\item Different perspectives. LBGAT mitigates natural accuracy degradation by focusing on the guidance of the natural classifier boundary.
Different from it, our proposed TRAIN emphasizes the importance of the topology of the sample in the representation space.
By combining the two perspectives, we can further enhance the model's performance, as confirmed by the experimental results.
\item Different interactions between models. In LBGAT, $M$ and ${M}'$ affect each other which still has the negative impact of the adversarial samples on the natural samples. 
However, when $M$ remains independent, optimizing LBGAT becomes challenging due to the inherent differences between the two models.
In TRAIN, $M$ unidirectionally influences ${M}'$ and as an anchor to preserve the original topology of natural samples in the representation space to avoid the negative influence of the adversarial samples on the natural samples.
This design choice effectively mitigates the adverse effects of adversarial samples on natural samples.
\end{enumerate}

These differentiating factors highlight the unique contributions of our proposed TRAIN method in addressing the natural accuracy degradation during adversarial training. By considering the topological aspects of samples to avoid the negative impact of adversarial samples, TRAIN offers a novel and effective approach for enhancing model robustness and performance.
\subsection{The Flexibility of TRAIN}

Different from other methods, TRAIN mitigates natural accuracy degradation by adopting a novel topological perspective.
Moreover, TRAIN could be applied to other adversarial training methods, such as vanilla AT~\citep{madry2017towards}, TRADES~\citep{zhang2019theoretically}, and LBGAT~\citep{cui2021learnable}, in a plug-and-play way.
To validate the effectiveness of our proposed enhancements, we conduct comprehensive validation experiments on these strong baselines.
We have introduced the robust loss of vanilla AT in our paper.
TRADES~\citep{zhang2019theoretically} improves classification performance by introducing a regularization term, which penalizes the discrepancy between the logits for adversarial examples and their corresponding natural images.
Its optimization objective is defined as:
\begin{equation}
\mathop{\arg\min} _{\theta} \mathbb{E}_{(x, y) \in D}\left(L(x,y;\theta)+\beta \max_{\delta \in S} {L}(x+\delta, x;\theta )\right),
\label{eq:formulate_trades}
\end{equation}
where $\beta$ is a hyper-parameter and its value means the strength of regularization for robustness.
TRADES has proven highly effective and remains a strong baseline for adversarial training to this day. 
Here we will elucidate the specifics of our approach when integrated with another strong baseline LBGAT.

LBGAT leverages the model logits obtained from a standard model to guide the learning process of an adversarial model. It is usually combined with vanilla AT and TRADES.
The total loss $L_{\rm AT}$ of adversarial model ${M}'$ combined with LBGAT and vanilla AT is:
\begin{equation}
\begin{split}
        L_{\rm {AT}} &= L({z}'_{{x}_{i}'} ,y_{i}) + \gamma ||{logit}'_{{x_{i}'}}-logit_{x_{i}}||_{2}\\
        &\!+ \!\lambda \! \sum_{i} \! \! \sum_{j}\! \!\left[p_{i| j}\! \log \!\left(\frac{p_{i| j}}{q_{i| j}}\!\right)\! +\!\left(\!1-p_{i |j}\!\right)\! \!\log\! \!\left(\!\frac{1-p_{i| j}}{1-q_{i| j}}\!\right)\!\right]\!,
\end{split}
    \label{eq:ours+lbgat+vanilla_at} 
\end{equation}
and the total loss of adversarial model $L_{\rm AT}$ combined with LBGAT and TRADES is:
\begin{equation}
    \begin{split}
        L_{\rm AT} &=L({z}'_{x_i} ,y_{i})
        +\beta KL({z}'_{x_i}||{z}'_{{x}'_i})
        + \gamma ||{logit}'_{{x'_{i}}}-logit_{x_{i}}||_{2}\\
        &\!+ \!\lambda \! \sum_{i} \! \! \sum_{j}\! \!\left[p_{i| j}\! \log \!\left(\frac{p_{i| j}}{q_{i| j}}\!\right)\! +\!\left(\!1-p_{i |j}\!\right)\! \!\log\! \!\left(\!\frac{1-p_{i| j}}{1-q_{i| j}}\!\right)\!\right]\!,\\
        &{z}'_{x_i} =\frac{exp({logit}'_{{x}_{i}})}{\sum_{j=1}^N exp({logit}'_{{x}_{j}})},
        {z}'_{{x'_i}} =\frac{exp({logit}'_{{x}'_{i}})}{\sum_{j=1}^N exp({logit}'_{{x}'_{j}})},
    \end{split}
    \label{eq:lbgat+trades} 
\end{equation}
where $KL$ is Kullback–Leibler divergence, which is commonly used to concretely implement the robust regularization of TRADES.
$\gamma$ is the hyper-parameter of LBGAT, and $\beta$ is the hyper-parameter of TRADES.

\subsection{Experimental Settings}
\label{experimental_settings}

\textbf{Datasets.}
Following~\citep{cui2021learnable,jia2022adversarial,pang2022robustness}, we conduct extensive evaluations on popular datasets, including CIFAR-10, CIFAR-100~\citep{krizhevsky2009learning} and Tiny ImageNet~\citep{deng2009imagenet} dataset to validate the effectiveness of our algorithm.
The CIFAR-10 and CIFAR-100 datasets consist of a total of $60,000$ color images with dimensions of $32 \times 32$ pixels.
Among these $50,000$ images are designated for training and the remaining $10,000$ images are reserved for testing.
Furthermore, CIFAR-10 has 10 categories while CIFAR-100 has 100 categories.
Furthermore, the Tiny ImageNet dataset includes $120,000$ color images with dimensions of $64 \times 64$ pixels in $200$ categories, with each category containing $500$ training images, $50$ validation images, and $50$ testing images.
This dataset offers a larger image size and a broader range of categories, enabling a more challenging evaluation of our algorithm's performance.

\textbf{Baselines.} We choose three strong baselines to demonstrate the effectiveness of our method: Vanilia AT~\citep{madry2017towards}, TRADES~\citep{zhang2019theoretically}, and LBGAT.
For TRADES, we set $\beta = 6.0$. 
For LBGAT, we conduct experiments based on vanilla AT and TRADES ($\beta = 6.0$).
We also add ALP~\citep{kannan2018adversarial} as a baseline in the Tiny ImageNet dataset.
In addition, we combine TRAIN with them to demonstrate the superiority of our approach.
To provide a comprehensive evaluation and comparison with other state-of-the-art adversarial training methods, we include additional baseline: MART~\citep{wang2019improving}, FAT~\citep{zhang2020attacks}, GAIRAT~\citep{zhang2021geometry}, AWP~\citep{wu2020adversarial}, SAT~\citep{sitawarin2021sat}, LAS~\citep{jia2022adversarial}, and ECAS~\citep{kuurila2023adaptive}.

\textbf{Evaluation metrics.}
To evaluate the generalization of the model on natural and adversarial samples, our evaluation metrics are natural data accuracy (Natural Acc.) and robust accuracy (Robust Acc.).
Robust accuracy is the model classification accuracy under adversarial attacks.
As specified in the respective publications, we choose three representative adversarial attack methods for evaluation: PGD-20, C\&W-20~\citep{carlini2017towards}, and Auto Attack~\citep{croce2020reliable}. 
We denote the model's defense success rate under those attacks separately as \textit{PGD-20 Acc.}, \textit{C\&W-20 Acc.}, and \textit{AA Acc.}.
Similar to manifold learning, we make $k$NN test accuracy as a topology score due to $k$NN relying solely on the relationships among samples to classify.
We utilize training sets as support sets (natural samples and adversarial samples generated by PGD-20) and methods~\citep{van2008visualizing,mcinnes2018umap}.
In Fig.~3 we set $k$ as $5,10,20,30,40,50$, and we observe that the choice of $k$ does not affect the relative ranking of the topological relationships among samples in different representation spaces. So in Tables~\ref{CIFAR-10} and~\ref{CIFAR-100}, we set $k$ as $30$.

\textbf{Data pre-process.}
Similar to LBGAT~\citep{cui2021learnable}, for CIFAR-10/100 datasets, the input size of each image is $32 \times 32$, and the training data is normalized to $[0, 1]$ after standard data augmentation: random crops of $4$ pixels padding size and random horizontal flip, and the test set is normalized to $[0, 1]$ without any extra augmentation;
For the training set of Tiny ImageNet, we resize the image from $64 \times 64$ to $32 \times 32$, and the data augmentation is random crops with $4$ pixels of padding; finally, we normalize pixel values to $[0,1]$,  and for the test set, we resize the image to $32 \times 32$ and normalize pixel values to $[0,1]$. Others are the same as CIFAR datasets.

\textbf{Training details.}
For Tables~1 and~2, we follow state-of-the-art adversarial training method LAS ~\citep{jia2022adversarial}.
$\epsilon$ is $8/255$, and 
The initial learning rate is set to $0.1$ with a total of 110 epochs for training and reduced to $0.1$x at the $100$-th and $105$-th epochs.
Weight decay is $5 \times 10^{-4}$, and random seed is $1$.
ResNet-18 is the backbone of standard models, and WideResNet-34-10 is the backbone of adversarial models.
The adopted adversarial attacking method during training is PGD-10, with a perturbation size $\epsilon = 0.031$, a step size of perturbations $\epsilon_1 = 0.007$. 
For different experiment settings, we choose different $\lambda$.
We set $\lambda = 5$ on CIFAR-10 dataset, and $\lambda = 20a$ on CIFAR-100 dataset, where $a = \frac{2}{1+ e^{-\frac{10t}{100}-1}}$ and $t$ is the current $t$-th epoch during training. 
Finally, all experiments were done on GeForce RTX 3090.

\subsection{Sensitivity of different learning rate}

\textbf{Experimental settings.}~For Table~\ref{CIFAR-10}, Table~\ref{CIFAR-100}, qualitative experiments, and all ablation experiments, we keep the same super-parameter configuration as LBGAT~\cite{cui2021learnable}.
The initial learning rate is set to $0.1$ with a total of $100$ epochs for training and reduced to $0.1$x at the $75$-th and $90$-th epochs. 
The optimization algorithm is SGD, with a momentum of $0.9$ and weight decay of $2 \times 10^{-4}$.
Moreover, all our experimental results are reproducible with a random seed of $1$.

Our method exhibits superior performance when applied with the new hyperparameters.
According to Tables~\ref{CIFAR-10} and~\ref{CIFAR-100}, TRAIN can effectively increase both natural and robust accuracy, and contribute to the topology preservation of both natural and adversarial samples.

In Table~\ref{CIFAR-10}, TRAIN gets an improvement by $2.16\%$ compared to vanilla AT baseline on natural data.
It surpasses vanilla AT on PGD-20, C\&W, and AA accuracy by $2.19\%$, $2.46\%$, and $1.94\%$ respectively, indicating its high robustness.
Our method also has improvements on LBGAT by $4.50\%$ to $1.87\%$ in all aspects.
For another common baseline, TRADES, TRAIN also gets competitive results on both natural and adversarial data. 
Note that natural accuracy decreases when applying LBGAT to TRADES, so it also brings a large enhancement when combined with our method.
For the topology score which is measured by $k$NN accuracy, TRAIN could boost the performance by a large margin.
Since $k$NN classification is based only on inter-sample relationships, such results prove that TRAIN could mitigate topology disruptions of both natural and adversarial samples from adversarial training.

The overall results on CIFAR-100 are similar to CIFAR-10. 
As shown in Table~\ref{CIFAR-100}, TRAIN performs better than vanilla AT and LBGAT and gets a further improvement when deployed with LBGAT simultaneously. 
For TRADES, our method surpasses it by a large margin ($8.78\%$) on natural data and improves the robust accuracy by $3.04\%$.
Adding LBGAT to TRAIN causes a decrease in natural accuracy but achieves the best accuracy in PGD-20 and C\&W-20.
The above results show that the proposed TRAIN could be applied to popular adversarial training pipelines for achieving SOTA performance on both natural accuracy and robust accuracy.
Despite a slight decrease in individual robust metrics, we have achieved a better balance between natural accuracy and adversarial robustness overall.
For the topology score, we can find that combining the baseline with TRAIN can further enhance the quality of topology for both natural and adversarial samples in the representation space.

\begin{table*}[]

    \centering 
    \scalebox{0.65}{
    \begin{tabular}{lcccccc}
    \toprule
    \multirow{2}{*}{\textbf{Defense}}
    &\multirow{2}{*}{\textbf{Natural Acc.}} &\multicolumn{3}{c}{\textbf{Robust Acc.}}
    &\multicolumn{2}{c}{\textbf{Topology Score}} \\
    ~&~&\textbf{PGD-20 Acc.}& \textbf{C\&W-20 Acc.}&\textbf{AA Acc.}&\textbf{Natural}&\textbf{Robust}\\
    \midrule 
    Standard Training & 94.46 & 0.00  & 0.00 & 0.00 & 94.94 & -\\
    \midrule 
Vanilla AT~\citep{madry2017towards}& 86.69 & 53.45  & 53.72 & 48.95 & 86.51 & 53.94\\
    Vanilla AT + TRAIN & \textbf{88.85}\textcolor{MocoGreen}{($\uparrow$ 2.16)}
    & \textbf{55.64}\textcolor{MocoGreen}{$(\uparrow 2.19) $}
    & \textbf{56.18}\textcolor{MocoGreen}{$(\uparrow 2.46) $}
    & \textbf{50.89}\textcolor{MocoGreen}{($\uparrow$ 1.94)}
    & \textbf{89.11}\textcolor{MocoGreen}{$(\uparrow 2.60)$} &\textbf{56.55}\textcolor{MocoGreen}{$(\uparrow 2.61)$} \\
     \cmidrule(r){1-7}
     Vanilla AT + LBGAT~\citep{cui2021learnable} & 86.55 & 54.34  & 53.35 & 47.27 & 86.64 & 54.26\\
     Vanilla AT + LBGAT + TRAIN & \textbf{89.42\textcolor{MocoGreen}{$(\uparrow 2.87)$}} & \textbf{56.21\textcolor{MocoGreen}{$(\uparrow 1.87)$}}  & \textbf{57.48\textcolor{MocoGreen}{$(\uparrow 4.13)$}} & \textbf{51.77\textcolor{MocoGreen}{$(\uparrow 4.50)$} }
     &\textbf{89.25}\textcolor{MocoGreen}{$(\uparrow 2.61)$}
     &\textbf{56.59}\textcolor{MocoGreen}{$(\uparrow 2.33)$}\\
     \cmidrule(r){1-7}
     TRADES*~\citep{zhang2019theoretically} & 84.42 & 56.59  & 54.91 & 51.91 & 85.58 & 56.73 \\
     TRADES + TRAIN & \textbf{87.30}\textcolor{MocoGreen}{$(\uparrow 2.88) $} & \textbf{58.20\textcolor{MocoGreen}{$(\uparrow 1.61) $}} & \textbf{56.31}\textcolor{MocoGreen}{$(\uparrow 1.40) $}& \textbf{53.09}\textcolor{MocoGreen}{$(\uparrow 1.18) $}
     & \textbf{90.01}\textcolor{MocoGreen}{$(\uparrow 4.43) $} & \textbf{58.86}\textcolor{MocoGreen}{$(\uparrow 2.13) $} \\
     \cmidrule(r){1-7}
     TRADES + LBGAT*~\citep{cui2021learnable} & 81.98 & \textbf{57.78}  & 55.53 & 53.14 & 84.57 & 57.79\\
     TRADES + LBGAT + TRAIN & \textbf{87.62}\textcolor{MocoGreen}{$(\uparrow 5.64) $}
     & 57.73\textcolor{red}{$(\downarrow 0.05)$}
     & \textbf{58.08\textcolor{MocoGreen}{$(\uparrow 2.55) $}}
     & \textbf{53.64\textcolor{MocoGreen}{$(\uparrow 0.50)$}}
     &\textbf{89.50}\textcolor{MocoGreen}{$(\uparrow 5.00)$} &57.98\textcolor{MocoGreen}{$(\uparrow 0.19)$} \\
\bottomrule
\end{tabular}}
\vspace{2mm}
\caption{Results on CIFAR-10.
When added to the existing baseline under most settings, our method achieves both natural accuracy and robust accuracy improvements, particularly in terms of C\&W-20 Acc.
``*'' are the results directly quoted from LBGAT.}
\label{CIFAR-10}
\end{table*}

\begin{table*}[]

    \centering 
    \scalebox{0.65}{
    \begin{tabular}{lcccccc}
    \toprule
    \multirow{2}{*}{\textbf{Defense}}
    &\multirow{2}{*}{\textbf{Natural Acc.}} &\multicolumn{3}{c}{\textbf{Robust Acc.}}
    &\multicolumn{2}{c}{\textbf{Topology Score}} \\
    ~&~&\textbf{PGD-20 Acc.}& \textbf{C\&W-20 Acc.}&\textbf{AA Acc.}&\textbf{Natural}&\textbf{Robust}\\
    \midrule 
    Standard Training & 77.39 & 0.00  & 0.00 & 0.00 &77.07 &- \\
    \midrule 
    Vanilla AT~\citep{madry2017towards}& 60.44 & 28.06  & 27.85 & 24.81 & 57.17 & 31.32\\
    Vanilla AT + TRAIN  & \textbf{66.39}\textcolor{MocoGreen}{$(\uparrow 5.95) $} & \textbf{29.88}\textcolor{MocoGreen}{$(\uparrow 1.82) $} & \textbf{29.84}\textcolor{MocoGreen}{$(\uparrow 1.99) $} & \textbf{25.81}\textcolor{MocoGreen}{$(\uparrow 1.00) $}
    &\textbf{64.70}\textcolor{MocoGreen}{$(\uparrow 7.53) $}
    & \textbf{32.84}\textcolor{MocoGreen}{$(\uparrow 1.52) $}
    \\
    \cmidrule(r){1-7}
    Vanilla AT + LBGAT~\citep{cui2021learnable} & 61.01 & \textbf{30.10} & 28.09 & 25.63 &61.28 &30.47 \\
      Vanilla AT + LBGAT + TRAIN & \textbf{68.20\textcolor{MocoGreen}{$(\uparrow 7.19) $}} & 29.83\textcolor{red}{$(\downarrow 0.27)$} & \textbf{30.84\textcolor{MocoGreen}{$(\uparrow 2.75) $}} & \textbf{25.88\textcolor{MocoGreen}{$(\uparrow 0.25) $}} & \textbf{66.08}\textcolor{MocoGreen}{$(\uparrow 4.80) $} &\textbf{32.48}\textcolor{MocoGreen}{$(\uparrow 2.01) $} \\
     \cmidrule(r){1-7}
     TRADES*~\citep{zhang2019theoretically} & 56.50 & 30.93  & 28.43 & 26.87 &52.57 & 32.17  \\
      TRADES + TRAIN & \textbf{65.28( \textcolor{MocoGreen}{$\uparrow$ 8.78}) } & \textbf{33.97\textcolor{MocoGreen}{$(\uparrow 3.04) $}}  & \textbf{30.86}\textcolor{MocoGreen}{$(\uparrow 2.43)$} & \textbf{28.25\textcolor{MocoGreen}{$(\uparrow 1.38) $}} & \textbf{65.78}\textcolor{MocoGreen}{$(\uparrow 13.21) $} & \textbf{34.53}\textcolor{MocoGreen}{$(\uparrow 2.36) $} \\
      \cmidrule(r){1-7}
      TRADES + LBGAT*~\citep{cui2021learnable} & 60.43 & 35.50  & 31.50 & \textbf{29.34} &61.06 &37.52\\
     TRADES + LBGAT + TRAIN & \textbf{62.62}\textcolor{MocoGreen}{$(\uparrow 2.19) $ }& \textbf{36.27\textcolor{MocoGreen}{$(\uparrow 0.77) $}}  & \textbf{31.72\textcolor{MocoGreen}{$(\uparrow 0.22) $}} & 29.19\textcolor{red}{$(\downarrow 0.15)$}& \textbf{64.84} \textcolor{MocoGreen}{$(\uparrow 3.78) $}&\textbf{38.25}\textcolor{MocoGreen}{$(\uparrow 0.73) $}\\
\bottomrule
\end{tabular}}
\vspace{2mm}
\caption{Results on CIFAR-100. Similar to Table~\ref{CIFAR-10}, our method can improve the natural accuracy (up to $8.78\%$), robust accuracy (up to $3.04\%$), and topology score (up to $13.21\%$) of baselines. ``*'' are the results directly quoted from LBGAT.}
\label{CIFAR-100}
\end{table*}
\subsection{Quantitative results on Tiny ImageNet.}
To demonstrate the effectiveness of our approach on a highly demanding dataset, we performed rigorous experiments on the Tiny Imagenet dataset.
The results, as depicted in Table~\ref{TinyImageNet}, clearly demonstrate that the combination of our algorithm with TRADES and LBGAT techniques leads to substantial improvements in both natural accuracy and adversarial robustness.
When combined with Trades, our approach achieves a $2.61\%$ improvement in natural accuracy and a $2.70\%$ improvement in robust accuracy. When combined with TRADES+LBGAT, our method achieves a $2.27\%$ improvement in natural accuracy and a $0.67\%$ improvement in robust accuracy.

\begin{table}[]
    
    \centering 
    \scalebox{0.5}{
    \begin{tabular}{lcc}
    \toprule
    \textbf{Defense} &
    \textbf{Clean Acc.} &\textbf{PGD-20 Acc.}\\
    \midrule 
    Vanilla AT*~\citep{madry2017towards} & 30.65 & 6.81 \\
    Vanilla AT + LBGAT*~\citep{cui2021learnable} &36.50 & 14.00\\
    ALP*~\citep{kannan2018adversarial} &30.51 &8.01\\
    LBGAT + ALP*~\citep{cui2021learnable} & 33.67 &14.55\\
    TRADES ($\beta$ = 6.0)*~\citep{zhang2019theoretically} & 38.51 & 13.48 \\
    TRADES ($\beta$ = 6.0) + LBGAT*~\citep{cui2021learnable} & 39.26 & 16.42\\
    \midrule
    TRADES ($\beta$ = 6.0) + Ours & 
    41.12\textcolor{MocoGreen}{$(\uparrow 2.61) $} & 16.18\textcolor{MocoGreen}{$(\uparrow 2.70) $}\\
    TRADES ($\beta$ = 6.0) + LBGAT+ Ours & \textbf{41.53}\textcolor{MocoGreen}{$(\uparrow 2.27) $} & \textbf{17.09}\textcolor{MocoGreen}{$(\uparrow 0.67) $} \\
\bottomrule
\end{tabular}}
\vspace{2mm}
\caption{Quantitative experiment on Tiny ImageNet. "*" are the results directly quoted from LBGAT. }
\label{TinyImageNet}
\end{table}

\subsection{More Ablation Studies}
In this section, we delve into TRAIN to study its effectiveness in batch size, hyper-parameter $\lambda$, and model architectures.
We also analyze the time complexity and training time of our method.
All the ablation experiments are based on the CIFAR-100 dataset and combined with TRADES.
All ablation experimental settings \textbf{(including ablation on different relationship preservation methods in our paper)} are the same as Tables~\ref{CIFAR-10} and~\ref{CIFAR-100}.

\begin{table*}[h]
   
    \centering 
    \scalebox{0.7}{
    \begin{tabular}{lllcccc}
    \toprule
    \multirow{2}{*}{\textbf{Backbone of $M'$}} & \multirow{2}{*}{\textbf{Training Strategy}}&\multirow{2}{*}{\textbf{Backbone of $M$}} &\multirow{2}{*}{\textbf{Clean Acc.}} &\multicolumn{3}{c}{\textbf{Robust Acc.}} \\
    ~&~&~&~&\textbf{PGD-20 Acc.}& \textbf{C\&W-20 Acc.}&\textbf{AA Acc.}\\
    \midrule 
    None & Standard Training & ResNet-18 & 77.39 & 0  & 0 & 0  \\
    \midrule 
    WideResNet34-10 & TRADES + Ours & ResNet-18 & 62.62 & 36.27  & 31.72 & 29.19  \\
    \midrule 
    None & Standard Training & WideResNet34-10 & 78.11 & 0  & 0 & 0  \\
    \midrule
    WideResNet34-10 & TRADES + Ours & WideResNet34-10 & 63.09 & 35.54  & 30.41 & 28.76  \\
\bottomrule
\end{tabular}}
\vspace{2mm}
 \caption{The ablation experiment about different backbones of the standard model.}
    \label{Ablation3}
\end{table*}

\begin{table}[h]
\centering
\resizebox{0.65\linewidth}{!}{
 \begin{tabular}{lccccc}
    \toprule
    \multirow{2}{*}{\textbf{Batch Size}} &\multirow{2}{*}{\textbf{Natural Acc.}} &\multicolumn{3}{c}{\textbf{Robust Acc.}} \\
    ~&~&\textbf{PGD-20 Acc.}& \textbf{C\&W-20 Acc.}&\textbf{AA Acc.}\\
    \midrule 
    128 & 66.39 & 29.88  & 29.84 & 25.81 \\
    256 & \textbf{66.55} & \textbf{31.08}  & \textbf{30.72} & \textbf{26.07} \\
    384 & 66.26 & 30.60  & 30.16 & 25.41 \\

\bottomrule
\end{tabular}
}
\vspace{2mm}
\caption{The ablation experiment about different batch sizes.}
\label{Ablation2}
\end{table}

\textbf{Impact of batch size.} As shown in Table~\ref{Ablation2}, we tried $128$, $256$, $384$ samples per batch for relation calculating. 
Among them, a batch size of $256$ achieves the best results, but the difference among different batch sizes is not large. Overall our method is not sensitive to different batch sizes. 

To ensure fair comparisons with other methods, we chose a batch size of 128 for our other experiments.

\begin{table}[!h]

    \centering 
    \scalebox{1.0}{
    \begin{tabular}{cccccc}
    \toprule
     &  0 & 5$a$ & 10$a$ & 20$a$  &  50$a$ \\
    \midrule 
     \textbf{Natural Acc.} & 57.99 & 61.52 & 63.21 & 65.28 & \textbf{66.40} \\
     \textbf{PGD-20 Acc.} & 31.53 & 32.31 & 33.47 & \textbf{33.90} & 33.62 \\
     \textbf{$L_{\rm TP}$} & 0.66 & 0.35 & 0.32 & 0.27 & \textbf{0.24}\\
\bottomrule 
\end{tabular}}
\vspace{2mm}
\caption{Sensitivity analysis of hyper-parameter $\lambda$.}
\label{tab:hyper}

\end{table}

\begin{table}[h]
    
    \centering 
    \scalebox{0.75}{
    \begin{tabular}{lcccc}
    \toprule
    \multirow{2}{*}{\textbf{Methods}} &\multirow{2}{*}{\textbf{Clean Acc}} &\multicolumn{3}{c}{\textbf{Robust Acc}} \\ 
    ~&~&\textbf{PGD-20 Acc}& \textbf{C\&W-20 Acc}&\textbf{AA Acc}\\
    \midrule 
    Vanilla AT & 60.44 & 28.06  & 27.85 & 24.81 \\
    Vanilla AT + TRAIN* & 65.15 & 28.00 & 27.90 & 24.91 \\
    Vanilla AT + TRAIN & \textbf{66.39} & \textbf{29.88} & \textbf{29.84}  & \textbf{25.81}\\
\bottomrule
\end{tabular}
}
\vspace{2mm}
\caption{Ablation experiment about different standard models on CIFAR-100. TRAIN$^{*}$ means using a well-trained standard model, and TRAIN means training two models jointly.}
    \label{Ablation:5}
\end{table}

\textbf{Sensitivity analysis of hyper-parameter $\lambda$.}
As Table~\ref{tab:hyper} shows, with the increase of $\lambda$ in Eq.~(3), natural accuracy always gets higher; $L_{\rm TP}$ (calculated from the test set) gets lower; while the PGD-20 accuracy rises at first and then remains stable.
It is reasonable because a large $\lambda$ forces the topology of clean samples to be highly close to that of standard models.
Finally, we set $\lambda$ as 20$a$ according to the PGD-20 accuracy following~\citep{madry2017towards}.

\textbf{Impact of the different standard models.}
As depicted in Table~\ref{Ablation3}, our approach exhibits robustness to variations in the backbones of standard models. Specifically, we observe that ResNet18 achieves a comparable trade-off between natural accuracy and adversarial robustness to WideresNet34-10 on the CIFAR-100 datasets while incurring lower training costs.

Table~\ref{Ablation:5} shows the results of using different standard training strategies.
To expedite the training process, a pre-trained standard model can be used in TRAIN (vanilla AT+TRAIN$^{*}$). However, training the standard model and adversarial model jointly achieves superior results.
This is attributed to the fact that the representation spaces of the two joint models are closer, facilitating optimization procedures.
\begin{table}[H]
    
    \centering 
    \scalebox{0.75}{
    \begin{tabular}{lcccc}
    \toprule
    \multirow{2}{*}{\textbf{Methods}} &\multirow{2}{*}{\textbf{Clean Acc}} &\multicolumn{3}{c}{\textbf{Robust Acc}} \\ 
    ~&~&\textbf{PGD-20 Acc}& \textbf{C\&W-20 Acc}&\textbf{AA Acc}\\
    \midrule 
    Vanilla AT & 35.10	&18.89&	16.19&	14.63\\
    Vanilla AT+ TRAIN &38.58	&20.25&	17.64	&15.39\\
    \midrule
    TRADES& 38.39&	17.90	&14.36&	13.38\\
    TRADES +TRAIN &43.64	&18.52&	14.86	&13.51\\
\bottomrule
\end{tabular}}
\vspace{2mm}
\caption{Experiments using MobileNetv3 on CIFAR100.}
    \label{Ablation:6}
\end{table}

\textbf{Impact of different backbones of $M'$.}
We conduct experiments on MobileNetv3, and the results reinforce the effectiveness of our approach across different backbones.
As shown in Table~\ref{Ablation:6}, our method can further improve the baseline, especially in natural accuracy.
We achieve a maximum improvement of $5.25\%$ in natural accuracy and a maximum improvement of $1.45\%$ in robust accuracy.

We can also find that the experimental results on MobileNet v3 are inferior compared to WideResNet34-10, both in terms of robustness and natural sample accuracy. This observation can be attributed to the positive correlation between the effectiveness of adversarial training algorithms and model capacity~\citep{bai2021recent}, and to reduce inference speed, MobileNet v3 has a significantly smaller model capacity compared to WideResNet34-10.

\bibliography{egbib}
\end{document}